\documentclass[11pt]{article}
\usepackage[final]{acl}

\usepackage{times}
\usepackage{latexsym}

\usepackage[T1]{fontenc}

\usepackage[utf8]{inputenc}
\usepackage{amsmath}
\usepackage{amsfonts}

\usepackage{microtype}

\usepackage{inconsolata}
\usepackage[table]{xcolor}

\usepackage{graphicx}

\usepackage[capitalize,noabbrev]{cleveref}

\usepackage{amsthm,mathbbol,amsmath,amssymb,amsfonts}

\usepackage{layouts}
\usepackage{url}
\usepackage{tabularx}
\usepackage{booktabs}       
\usepackage{siunitx}        
\usepackage{lineno}
\usepackage{makecell}
\usepackage{textcomp} 
\usepackage{wrapfig}
\usepackage{caption}
\usepackage{subcaption}
\usepackage{mwe} 
\usepackage[table,dvipsnames]{xcolor}
\usepackage{xcolor,colortbl}
\usepackage[most]{tcolorbox}
\usepackage{multirow}
\usepackage{arydshln}
\usepackage{stackengine}  
\usepackage{xspace}
\usepackage{soul} 
\usepackage[export]{adjustbox}

\usepackage{algorithm}
\usepackage{algorithmic}
\usepackage{caption}

\usepackage{listings}
\usepackage{fvextra}
\newcommand{\postspace}{\vskip -3mm}

\newcommand{\eg}[0]{\textit{e.g.},\xspace}
\newcommand{\ie}[0]{\textit{i.e.},\xspace}
\newcommand{\spicedcontrolless}{\texttt{Less-SciFi}\xspace}
\newcommand{\spicedcontrolmore}{\texttt{More-SciFi}\xspace}
\newcommand{\spicedcontrolsame}{\texttt{Same-SciFi}\xspace}
\newcommand{\gptbig}{\texttt{GPT-5.1}\xspace}
\newcommand{\geminibig}{\texttt{Gemini 3.1 Pro (Preview)}\xspace}

\newcommand{\gengpt}{\texttt{gpt-5.4-nano}\xspace}
\newcommand{\gengemini}{\texttt{Gemini 3.1 (fl)}\xspace}
\newcommand{\genllama}{\texttt{Llama 3.3 (70B)}\xspace}
\newcommand{\genqwen}{\texttt{Qwen3 (8B)}\xspace}
\newcommand{\genclaude}{\texttt{Claude 4.5 (H)}\xspace}

\newcommand{\evalgpt}{\texttt{gpt-5.4-mini}\xspace}

\newcommand{\certaintydistortion}{CD\xspace}
\newcommand{\certincrease}{CD-$\uparrow$\xspace}
\newcommand{\certdecrease}{CD-$\downarrow$\xspace}

\definecolor{bestcell}{RGB}{198, 239, 206}   
\definecolor{sectionbg}{RGB}{242, 242, 242}  
\definecolor{headergreen}{HTML}{2D5A45}

\title{
From `May' to `Is': Certainty Distortion in Language Model Rewriting
}
\author{
 {\bf Catarina G Belem\textsuperscript{1}},
 {\bf Shang Wu\textsuperscript{1}},  
 {\bf Hongyu Yao\textsuperscript{3}},\\
 {\bf Mark Steyvers\textsuperscript{1,2}},
 {\bf Sameer Singh\textsuperscript{1}},
 {\bf Padhraic Smyth\textsuperscript{1}}
\\
\\
 \textsuperscript{1}Department of Computer Science, University of California Irvine\\
 \textsuperscript{2}Department of Cognitive Science, University of California Irvine\\
 \textsuperscript{3}Massachusetts Institute of Technology\\
\\
 \small{
   \textbf{Correspondence:} \href{mailto:cbelem@uci.edu}{cbelem@uci.edu}
 }
}

\begin{document}
\maketitle
\begin{abstract}
Humans increasingly turn to Language Models (LMs) in ways that shape beliefs and drive decisions, including discussing, rewriting, and summarizing information from scientific articles, news, and medical reports. 
However, in these domains, where it often matters how confidently a claim is expressed, little is known about whether LMs faithfully preserve the degree of confidence. 
In this work, we investigate \textit{certainty distortion} in LMs, defined as meaningful changes in expressed certainty during transformations intended to preserve meaning. 
We propose an LM-based evaluation metric that is consistent with population-level judgments of certainty.
Using this metric, we characterize certainty distortion across different sizes and families of models in the context of scientific and medical communication tasks.
Our results show that certainty distortion affects up to 75\% of LM outputs 
and is systematically asymmetric in rewriting tasks with most LMs being 1.5-2$\times$ more likely to increase the expressed certainty than to decrease it. 
These effects can compound over repeated paraphrasing: in the medical domain, \genclaude increases certainty in 20\% of examples after a single iteration, increasing to 40\% after five iterations. 
Prompt-based interventions reduce overall certainty distortion but do not eliminate it.
Together, these findings reveal a general bias toward inflating expressed certainty, with direct implications for users who rely on LMs in high-stakes domains.
\end{abstract}
\section{Introduction}
\label{sec:introduction}
The use of language models (LMs) for summarizing, simplifying, and rewriting complex information is rapidly growing. In domains like medicine, law, and science, claims are carefully worded to reflect the degree of certainty supported by the evidence~\citep{BioScope-Vincze2008,Lee2017,Shinagare2019,Fore2019}. Authors use epistemic markers, such as hedges (\eg \textit{may}, \textit{could}), non-factive verbs (\eg \textit{suggest}, \textit{appear}), source attribution (\eg \textit{according to}), to signal what is established versus what is tentative~\citep{Fischhoff2014,dhami2022communicating}. When these markers are modified, the epistemic status of a claim shifts, and readers may form beliefs or make decisions that go beyond what the evidence supports~\citep{wu2025airesponsesshapeuser,guingrich2026beliefoffloadinghumanaiinteraction}.
\begin{figure}[tb]
    \centering
    \includegraphics[width=0.9\linewidth]{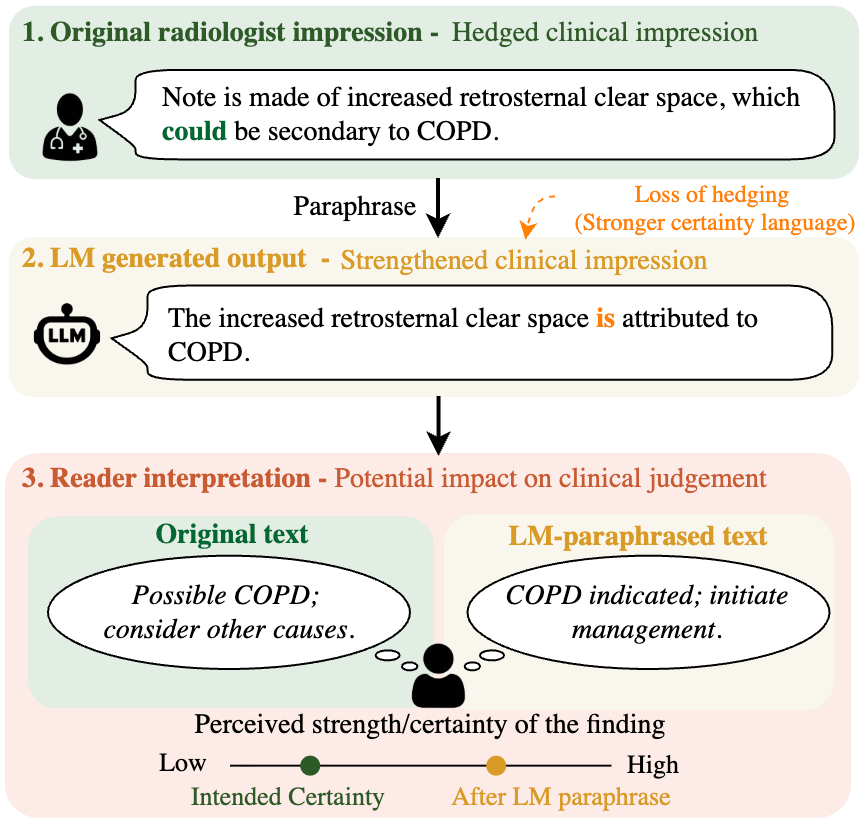}
    \caption{
    A real example of certainty inflation: \genqwen paraphrases a hedged radiology impression into a confident attribution, shifting the reader's belief from a tentative diagnostic to a definite clinical conclusion.
    }
    \label{fig:motivating-example}
\end{figure}
Prior work suggests that around 30\% of human-written scientific findings are inflated when communicated across media outlets~\citep{Brechman2009,Sumner2016,wuehrl-etal-2024-understanding}, indicating that certainty distortion is a real phenomenon in human-mediated communication. 
However, whether LMs exhibit similar behavior in rewriting tasks, and to what extent this may be a systematic aspect of LM behavior, remains an open question. 
Consider the motivating example in \cref{fig:motivating-example} concerning a radiology report from MIMIC-CXR~\citep{Johnson2019-MIMIC-CXR}:
the report describes the increased empty space behind the sternum as possibly explained by a chronic obstructive pulmonary disease (COPD), among other diagnoses. The LM paraphrase drops the hedge, attributing the empty space to COPD. Such a shift, if relied on clinically, could 
prematurely push the clinician toward treating COPD.

In this work, we investigate \textit{certainty distortion} in LMs.\footnote{All data and code are publicly available at \url{https://github.com/UCIDataLab/certainty-distortion-in-llm-rewriting}.}
We define certainty distortion as the likelihood that the degree of certainty expressed in source text is not preserved after an LM rephrases it.
We make three contributions.
First, we propose an LM-based evaluation metric for measuring certainty distortion, which we validate via a user study with multiple human annotations. 
We find that the proposed LM evaluation metric aligns with population-level consensus judgments as well as any individual human annotator does.
Second, we use this metric to systematically characterize certainty distortion across model sizes and families for scientific communication and medical communication tasks. 
We find that certainty distortion affects up to 75\% of LM outputs and that it is asymmetric, with most models being 1.5-2$\times$ 
more likely to amplify rather than attenuate expressed confidence.
When evaluating across repeated LM rewriting, we observe a contrast: in scientific communication tasks, most certainty shifts occur after the first LM output, whereas in medical communication tasks certainty shifts compound across iterations.
Specifically, in the medical communication task, we observe that models’ tendency to increase expressed certainty grows by an additional 13-20\% points by the fifth paraphrase.
Third, we evaluate standard mitigation strategies and find that greedy decoding consistently increases certainty distortion across models, primarily by amplifying shifts toward greater expressed certainty, whereas prompt-based interventions can reduce distortion but do not eliminate it.
    
Together, these findings show that the evaluated LMs systematically distort expressed certainty when rewriting text. Most exhibit a recurring tendency toward certainty inflation, which can compound across repeated medical paraphrases in particular.
Combined with growing evidence that more confident model outputs drive stronger belief updates~\citep{wu2025airesponsesshapeuser,guingrich2026beliefoffloadinghumanaiinteraction} and overreliance in users~\citep{Kim-et-al-2024-overreliance,Steyvers2025-NMI}, this points to a real risk in everyday LM use, one that current mitigation strategies fail to address. 
We argue that \textit{epistemic faithfulness}, \ie preserving not just what is said, but the degree of certainty with which it is said, deserves attention alongside factuality.

\section{Related Work}
\label{sec:related_work}
\paragraph{Certainty Distortion in Human Communication.}
Epistemic markers (\eg hedges, plausibility shields and non-factive verbs, evidential markers) encode a claim's evidential status and the speaker's commitment to it, distinguishing, for example, \textit{she might have committed fraud} from \textit{she definitely committed fraud}~\citep{zhou-etal-2023-navigating,zhou-etal-2024-relying}. 
Social science research further shows that epistemic markers (\eg \textit{likely}, \textit{doubtful}) correspond to measurable differences in inferred certainty ~\citep{Budescu1988,wallsten2008intel}.
Studies of scientific dissemination through press releases, news, and social media find that such epistemic markers are not faithfully preserved~\citep{Brechman2009,Sumner2016,wuehrl-etal-2024-understanding}.
However, these studies focus on communication across human intermediaries and do not establish whether LMs exhibit analogous patterns.

\paragraph{\textbf{Information Distortion in LM Transformations.}}
Recent work raises concerns that LMs can distort human-generated language, including by altering textual meaning~\citep{abdulhai2026llmsdistortwrittenlanguage}. 
Prior work on LM transformations primarily evaluates whether source semantic meaning and factual claims are preserved~\citep{maynez-etal-2020-faithfulness,pagnoni-etal-2021-understanding,Wang-survey-factuality,ramprasad2025do-factuality}.
Closer to our setting, recent work adopts the \textit{broken telephone} paradigm to study cumulative distortion:
\citet{perez2025when} measures drift in surface attributes, such as toxicity, sentiment, and length under iterative paraphrasing and open-ended generation, 
while \citet{mohamed-etal-2025-llm} examines factual and semantic drift in translation chains.
Both examine drift along axes distinct from epistemic certainty, focusing on factual or semantic preservation rather than shifts in expressed certainty during rewriting intended to preserve meaning. 
We treat expressed certainty as a separate axis
from factuality (whether a claim is true) 
and from calibration {
(whether a model's predicted probabilities or confidence estimates are statistically aligned with the empirical frequency of correctness)}~\citep{kadavath2022languagemodelsmostlyknow} 
and further show that this drift is directional, with a systematic bias toward increasing certainty.

\paragraph{Systematic Biases in LM Outputs.}
Prior work documents biases such as sycophancy, where models align responses with user beliefs rather than evidence~\citep{perez-etal-2023-discovering,sharma2024towards-sycophancy},
and brittleness to epistemic markers, which can sway LM discriminative judgments of accuracy and misinformation~\citep{lee-etal-2025-llm,vandesande2025factfrequencyllmresponses}. 
We extend this literature to generative tasks that require preserving source certainty.

\section{Measuring Certainty Differences}
\label{sec:methodology}
In this paper, we focus on information-preserving tasks like rephrasing and simplification. 
In these tasks, an LM is given a source text $x_o$ and is asked to generate an output $x_M$ that preserves all or part of the information in the source text. 
We are particularly interested in whether the \textit{level of certainty} in the input text remains the same in the transformed output text generated by the model. 
Here ``level of certainty'' is associated with semantic propositions being expressed in the text. 
For example, the text ``From our experiments, there is evidence to suggest that X may cause Y'' contains some uncertainty about the relationship between X and Y, whereas the text ``From our experiments we conclude that X causes Y'' contains no uncertainty, despite preserving the same semantic meaning. 
To characterize such \textit{certainty differences}, we consider text pairs: the original text $x_o$ and its model-generated version $x_{M}$.
Assuming a function $\text{c}(x) \in [0, 1]$ that reliably estimates the expressed certainty of a text, 
we define \textit{per-example certainty difference} as
\begin{equation}
\delta(x_o, x_M)  = \text{c}(x_M) - \text{c}(x_o),
\end{equation}
where 
$\delta = 0$ indicates preservation of certainty (no distortion), 
$\delta> 0$ indicates increased certainty (\textit{overconfidence}), 
and $\delta < 0$ indicates decreased certainty (\textit{underconfidence}).

To assess whether certainty distortion is a systematic property of model behavior rather than an artifact of individual examples, 
we define \textbf{Certainty Distortion (CD)} as the fraction of examples in a dataset $\mathcal{D}$ in which a shift occurs:
\begin{equation}
    \text{CD}(\mathcal{D}) = \mathbb{E}_{(x_o, x_M)\sim \mathcal{D}}[ \mathbb{1}_{[\delta(x_o, x_M) \neq 0]}].
\end{equation}

Additionally, we can further decompose \certaintydistortion into upward (\certincrease) and downward (\certdecrease) shifts to capture the distortion direction, where we track $\mathbb{1}_{[\delta(x_o, x_M) > 0]}$ and $\mathbb{1}_{[\delta(x_o, x_M) < 0]}$, respectively.

\subsection{Operationalizing Certainty Estimates}
\label{ssec:operationalizing-certainty-estimate}
%
%
%
%
To operationalize CD, we require a reliable estimator $\delta(x_o, x_M)$. 
One approach is to estimate certainty  $c(x)$ independently for each text $x$ and then compute the difference. 
However, this \textit{reference-less} approach could fail to capture subtle directional shifts in certainty, which could render it unreliable for estimating CD.
Instead, we treat $c(x)$ as a latent quantity and estimate $\delta$ directly through pairwise comparison. 

Human annotations of pairwise certainty differences would be ideal, but they are inherently noisy and subjective for this type of task~\citep{pei-jurgens-2021-measuring,wuehrl-etal-2024-understanding}. As a result, a large number of human annotations would be necessary to obtain reliable estimates, which is not scalable.

As an alternative, prior work has explored automated proxies based on fine-tuned BERT models~\citep{pei-jurgens-2021-measuring,wuehrl-etal-2024-understanding}. 
However, because these models operate in a reference-less manner, we hypothesize that they have limited sensitivity to subtle pairwise differences in certainty. 
We validate this hypothesis empirically by constructing controlled datasets in which certainty shifts are systematically manipulated (Appendix \ref{apdx:controlled-eval}).
We then evaluate whether these metrics align with the intended shift across subsets, finding that they fail to reliably detect. 
In fact, we observe a saturation effect where the BERT-based metrics assign uniformly high scores, leaving little headroom to distinguish moderately certain from highly certain texts. 
We further corroborate these findings by analyzing cases where the models disagree (\Cref{tab:qualitative-pairwise-examples:pei-jurgens-lm-judge,tab:qualitative-pairwise-examples:wuerl-et-al-lm-judge}). 

With these limitations in mind, in this work we pursue a pairwise evaluation approach that can directly compare two texts without relying on absolute scores. 
Specifically, we propose to estimate \certaintydistortion directly through an LLM-as-judge approach~\citep{liu-etal-2023-g}. 
We ask \evalgpt to determine which of two texts communicates the main finding with greater certainty. 
To avoid biases associated with the use of labels such as \textit{Original} and \textit{Modified}, we present the two texts $x_o$ and $x_M$ blindly as \textit{Text A} and \textit{Text B}, and prompt the model to select from a 5-point scale ranging from \textit{Clearly A} to \textit{Clearly B}, with \textit{No Clear Difference} as the midpoint (full prompt in \Cref{fig:prompt:llm-judge-5-options-structured}).\footnote{
The scale is ordinal rather than calibrated thresholds: its labels represent ordered differences.
We validate it for source--transformation pairs expressing the same underlying claim (Appendix \ref{apdx:automatic-metric-validation}). Other uses would require separate validation.
}
To encourage the model to explicitly formalize a certainty estimate for each text (\ie $c(x_o)$, $c(x_M)$) prior to making a comparative judgment, we elicit chain-of-thought (CoT) reasoning~\citep{Wei-cot-2022} over the certainty sources present in each text.\footnote{The improved alignment of using CoT with certainty shifts is validated in a controlled setting in Appendix \ref{apdx:controlled-eval}.} 
To further marginalize over positional bias~\citep{zhao2021calibrate,stureborg2024large,pezeshkpour-hruschka-2024-ordering}, we query the model twice reversing the input order, remapping each prediction into a canonical form--\textit{Clearly Original}, \textit{Slightly Original}, \textit{No Clear Difference}, \textit{Slightly Modified}, or \textit{Clearly Modified}--accounting for the input order of each call. 
When both calls produce the same canonical label, that label is adopted as the final prediction; when they agree in direction but differ in magnitude, we select the label closer to \textit{No Clear Difference}.
When they disagree in direction, 
we flag the pair as order-inconsistent and treat it as a tie when computing directional-shift rates. A moderator-based analysis indicates that most order-inconsistent pairs are best characterized as ties (Appendix \ref{apdx:additional-method-details-llm-judge}).
For analysis, we map the canonical labels to a numerical scale $\{-2,-1,0,1,2\}$, where negative values indicate greater certainty in the original text $x_o$, positive values indicate greater certainty in the LM-generated text $x_M$, and 0 indicates no perceived difference.
%

\subsection{Human Validation}
\label{ssec:human-validation}

To verify that \evalgpt serves as a reliable proxy for human certainty judgments, we conduct a human annotation study comparing LLM-as-a-judge predictions against human pairwise assessments of certainty. 
Specifically, we select 240 text pairs from the paraphrasing and news tasks in \Cref{sec:experiments}. 
We recruited 138 annotators from Prolific.\footnote{\url{https://www.prolific.com/}} 
After applying pre-specified quality filters, we excluded 9 annotators (3 for excessive off-screen time, 1 for fast uniform responses, and 5 for failing the attention check) leading to 129 total annotators.\footnote{Compliant with IRB protocols; see Appendix \ref{appdx:sec:details-user-studies}.}

As in the LLM-as-a-judge setup, the two texts $x_o$ and $x_M$ are presented anonymously as \textit{Text A} and \textit{Text B} in randomized order. Annotators judge which text communicates greater certainty on a 5-point Likert scale, subsequently remapped to the same numerical scale $\{-2,-1,0,1,2\}$ introduced in \Cref{ssec:operationalizing-certainty-estimate}, facilitating direct comparison between human and LLM-as-a-judge predictions.

We measure inter-annotator agreement using Krippendorff's $\alpha$, finding $\alpha \approx 0.25$ across annotator groups. 
The observed agreement is consistent with prior work on certainty annotation~\citep{rubin-2007-stating,wuehrl-etal-2024-understanding}, reflecting the inherent subjectivity of certainty judgments. 

\paragraph{Results.} 
To assess whether \evalgpt can serve as a reliable proxy for human judgment, we evaluate it in terms of its correlation with population-level perceptions of certainty. 
Specifically, we compute Kendall $\tau_B$ correlations between each annotator (and \evalgpt) and the \textit{human consensus}, defined via leave-one-out as the mean prediction of the remaining annotators.
If \evalgpt correlates more strongly with the consensus than the average annotator, it is a more reliable proxy for the central tendency of human certainty perception than any single annotator.  
As a complementary metric, we also report in  Appendix \ref{appdx:sec:details-user-studies}, the average consistency between \evalgpt and \textit{individual} human annotators, defined as the pairwise rank correlations between all \texttt{Human-LM} and \texttt{Human-Human} annotator pairs. 
If \evalgpt agrees with the individual humans at least as well as two humans agree with each other, it can be considered a reliable substitute for a randomly sampled annotator. 
As reference points, we include a random baseline and the 3-class classifier from \citet{wuehrl-etal-2024-understanding}.

\begin{figure}[tb]
    \centering
    \includegraphics[width=0.90\linewidth]{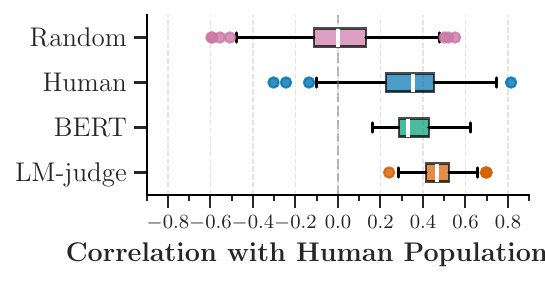}
    \caption{
    Correlation of different \certaintydistortion estimators with the mean prediction of human-population, measured using Kendall $\tau_B$ correlations. 
    Overall, the LM-judge correlates better with human population than the average individual Human or the strongest BERT-based estimator~\citep{wuehrl-etal-2024-understanding}.
    }
    \label{fig:human-validation:spiced-paraphrase-population-level-consensus-only}
    \postspace
\end{figure}

\Cref{fig:human-validation:spiced-paraphrase-population-level-consensus-only} 
shows that \evalgpt correlates with the human consensus more strongly than the average individual annotator ($\tau_B = 0.47_{\sigma = 0.09}$ vs $\tau_B=0.34_{\sigma =0.19}$), 
indicating that it captures the central tendency of human certainty perception at least as well as a randomly sampled annotator. 
At the individual-annotator level, \evalgpt also achieves consistency comparable to Human-Human consistency (Appendix \ref{appdx:sec:details-user-studies}), confirming its reliability as a substitute annotator.
 
Compared with the prior BERT-based estimator, \evalgpt correlates more strongly with the human consensus. Across 10,000 paired bootstrap samples, the mean correlation gap was $+0.16$ in favor of \evalgpt (95\% CI $[0.07, 0.25]$). In addition, \evalgpt provides a textual rationale alongside each judgment, facilitating qualitative inspection and debugging of how certainty differences are identified, and does not require supervised training on a task-specific data distribution. We therefore use \evalgpt as our certainty estimator throughout the remainder of the paper.
\section{Experiments}
\label{sec:experiments}

We now investigate the degrees to which LMs preserve or distort the epistemic certainty of original content.
To characterize this behavior, we report the fraction of examples in which a model-generated output differs in certainty from the source (\certaintydistortion), 
and further decompose this distortion into upward (\certincrease) and downward (\certdecrease) shifts to capture the distortion direction.

\subsection{Experiment Setup}
\label{ssec:experiment-setup-single-sent-level}

\paragraph{Datasets.} 
We begin with sentence-level tasks as our first testbed. 
At this granularity, models are less likely to introduce hallucinated or unrelated content~\citep{pmlr-v235-zhang24ay,zhao-etal-2025-response,yang2026hallucinatelongresponsegeneration}, 
and, as a result, \certaintydistortion is more likely to faithfully capture changes in epistemic certainty rather than differences in content.
We use two datasets: 
SPICED~\citep{wright-etal-2022-SPICED} and MIMIC-CXR~\citep{Johnson2019-MIMIC-CXR}.
SPICED is a dataset of scientific findings paired with corresponding news and social media versions, spanning various disciplines (\eg biology, computer science, and psychology). 
We focus specifically on the paired \textit{News} subset and filter out examples in which the original finding spans multiple sentences, resulting in 397 sentence pairs.  
Additionally, we use the Findings section of the radiology reports available in MIMIC-CXR. We restrict the analysis to a subset of 800 sentences across reports, 400 of which are selected as containing at least one epistemic marker (\eg \textit{suggests}, \textit{compatible with}). Additional details are available in Appendix \ref{apdx:sec:additional-dataset-details}.

\paragraph{Tasks.} 
We evaluate LMs on two tasks within each domain. In the scientific domain, we instruct models either to paraphrase a scientific finding or to rewrite it as if reporting it in The New York Times. In the medical domain, we similarly instruct models either to paraphrase a sentence from the findings section of a radiology report or to rewrite it in simplified language for a non-expert patient. All generation prompts are provided in Appendix \ref{apdx:prompts}. In this initial study, we intentionally keep prompts simple to better reflect how everyday users are likely to interact with these systems.

\paragraph{Models.}
We evaluate five instruction-tuned LMs from different model families that are widely used in practice: \gengemini~\citep{google2026gemini31pro}, \gengpt~\citep{openai_gpt54_2026}, and \genclaude~\citep{anthropic2024claudehaiku} as representative closed-source models, and  \genllama~\citep{grattafiori2024llama3herdmodels}, and \genqwen~\citep{yang2025qwen3technicalreport} as representative open-source models.
Except when explicitly stated otherwise, all experiments use top-p sampling (\texttt{T=1}, \texttt{top\_p=0.9}), reflecting how LMs are commonly used in practice~\citep{Holtzman2020The,li2023contrastive,scholten2025probabilistic}. 
Across all experiments, models are given a generation budget of 3.5k tokens. 
Finally, we evaluate the \certaintydistortion metric for each pair of original and generated texts, using \evalgpt as the LLM-as-a-judge, as described in \Cref{sec:methodology}.

\subsection{Certainty Distortion Results}
\label{ssec:sentence-level-distortion-rates}

\emph{\textbf{LMs exhibit non-negligible certainty distortion rates across tasks and domains}}.
\Cref{tab:cd-only:spiced-and-medical} summarizes the \certaintydistortion metric across tasks. 
Average \certaintydistortion rates range from 30\% for \gengpt to 49.4\% for \genllama.
Even in sentence-level paraphrasing of scientific findings, a task with minimal incentive for models to alter content or length, \emph{37.5\% to 64\% of LM-generated outputs distort the source text's expressed certainty}.
%

\emph{\textbf{News rewriting amplifies certainty distortion more than paraphrasing}}.
In the scientific domain, news rewriting pushes distortion up to 74.8\% (\genclaude) and 75.3\% (\gengemini), increases of 26.7 and 15.1 points respectively over paraphrasing for the same models. 
The difference is smaller for \gengpt and \genllama, with only a 5.3 to 6.8-point increase. 
Manual inspection (see Appendix~\ref{tab:qualitative-pairwise-examples:random:llms}) shows that 
\genclaude and \gengemini adopt the evidential framing typical of science journalism (\eg ``Researchers have'', ``Scientists have'') and stronger verbs (\eg ``discovered''), both of which strip hedges and present results as established.\footnote{
The use of journalistic framing suggests that genre conventions may partly contribute to the higher distortion observed in this task. As a diagnostic magnitude check, we apply the same evaluator to the corresponding human-written SPICED news texts, which exhibit 23.9\% CD, compared with 44.3--75.3\% for the LM rewrites. This is not a controlled human--LM comparison, as the human texts were naturally occurring and may reflect additional context, editorial goals, and other genre conventions. Nevertheless, the comparison suggests that LM distortion is not explained solely by reproducing the average distortion observed in human-written science news (Appendix \ref{ssec:human-spiced-analysis}). Moreover, genre conventions cannot explain the distortion observed in non-news tasks, including scientific and medical paraphrasing.
}
Conversely, \gengpt mostly preserves the source's voice (``We find'', ``Our results show'') and the same verbs, leaving less room for confidence shifts to occur. 
{
Since omitting hedges may also shorten outputs, these qualitative patterns raise the possibility that the observed inflation partly reflects compression.
We investigate this possibility in Appendix \ref{apdx:sec:length-analysis}, where regression controls and a matched-length analysis show that differences in length do not fully explain the effect.
}

\emph{\textbf{Certainty distortion is lower in medical communication than in scientific communication}}.
While distortion rates range between 37.5\%-75\% in scientific communication, the distortion rates in medical domain are much lower, 12.5\%-38.3\%.
One possible explanation lies in how each type of content relates to the model's prior knowledge. 
Scientific findings discuss general claims about the world that LMs may have encountered during pretraining, which may lead them to pattern-match to familiar phrasings rather than preserve the source's hedging. 
Radiology findings, by contrast, are observations about a specific patient that the model has no prior information, potentially leaving less room for the model to substitute more certain formulations learned from similar texts during pretraining.
\begin{table}[t]
\small
\centering
\setlength{\tabcolsep}{3pt}  
\begin{tabular}{l S[table-format=2.1] S[table-format=2.1]  S[table-format=2.1] S[table-format=2.1]  S[table-format=2.1]}
\toprule
& \multicolumn{2}{c}{{\textbf{SPICED}}} & \multicolumn{2}{c}{{\textbf{MIMIC}}} \\
\cmidrule(lr){2-3} \cmidrule(lr){4-5}
\textbf{Method} & \textbf{Para} & \textbf{News} & \textbf{Para} & \textbf{Simpl} &  \textbf{Avg} \\
\midrule
\gengpt         & 37.5 & 44.3 & 25.8 & 12.5 & 30.0 \\
\genclaude      & 48.1 & 74.8 & 29.1 & 20.1 & 43.0 \\
\gengemini      & 60.2 & 75.3 & 24.3 & 26.8 & 46.7 \\
\genqwen        & 43.1 & 57.7 & 31.1 & 18.2 & 37.5\\
\genllama       & 64.0 & 69.3 & 38.3 & 26.0 & 49.4 \\
\bottomrule
\end{tabular}
\caption{Model certainty distortion (in \%) on paraphrasing and rewriting of single-sentence scientific and radiological findings. Lower is better.  All 95\% CIs $\leq 0.1$.}
\label{tab:cd-only:spiced-and-medical}
\postspace
\end{table}

\emph{\textbf{Most models inflate certainty more than they deflate it}}.
\Cref{tab:cd-directional-ratio} decomposes certainty distortion into upward (CD-$\uparrow$) and downward (CD-$\downarrow$) shifts. 
Across both domains, most models inflate the certainty of the source more often than they deflate it, with inflation-to-deflation ratios ranging from 1.5 to 2.0. 
We note two exceptions: \gengpt and \genllama are roughly balanced across SPICED tasks (0.9 and 1.0, respectively), 
    and \genllama further stands out on MIMIC, where it is the only model to significantly deflate more than it inflates (0.6).
\begin{table}[tb]
\centering
\small
\setlength{\tabcolsep}{3pt}
\begin{tabular}{lccc ccc}
\toprule
& \multicolumn{3}{c}{\textbf{SPICED}} & \multicolumn{3}{c}{\textbf{MIMIC}} \\
\cmidrule(lr){2-4} \cmidrule(lr){5-7}
\textbf{Model} & \textbf{\certincrease} & \textbf{\certdecrease}  & \textbf{R} & \textbf{\certincrease}  & \textbf{\certdecrease} & \textbf{R} \\
\midrule
\gengpt     & 19.0 & 21.9 & 0.9          & 14.4 &  4.8 & 3.0 \\
\genclaude  & 37.5 & 24.2 & 1.5          & 16.0 &  8.6 & 1.9 \\
\gengemini  & 42.9 & 24.8 & 1.7          & 15.5 & 10.1 & 1.5 \\
\genqwen    & 33.4 & 17.0 & 2.0          & 16.1 &  8.6 & 1.9 \\
\genllama   & 33.5 & 33.1 & 1.0          & 12.3 & 19.9 & 0.6 \\
\bottomrule
\end{tabular}
\caption{
Directional certainty distortion rates (\%) and inflation-to-deflation ratio, averaged across tasks within each domain. 
Ratios R $\geq$ 1 indicate models inflate certainty more often than they deflate it. 
}
\label{tab:cd-directional-ratio}
\postspace
\end{table}

\begin{figure}[tb]
    \centering
    \includegraphics[width=\linewidth]{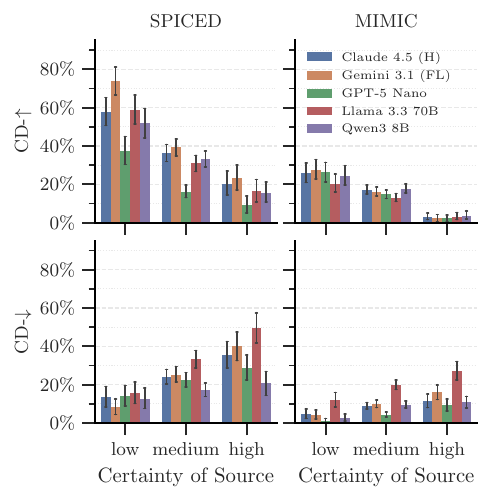} 
    \caption{
    Direction of certainty distortion (CD-$\uparrow$ top, CD-$\downarrow$ bottom) as a function of the source's certainty level (low/medium/high) across SPICED (left) and MIMIC (right). 
    As expected from floor/ceiling effects, CD-$\uparrow$ decreases and CD-$\downarrow$ increases with source certainty. 
    However, the magnitudes are asymmetric: across most models and both domains, inflation of uncertain sources is more frequent than deflation of certain sources. 
    Error bars show 95\% CI.}
    \label{fig:cd-by-source-certainty}
    \postspace
\end{figure}

\Cref{fig:cd-by-source-certainty} breaks the analysis down by the expressed certainty of the original text.  
Although certainty distortion varies with source certainty,
we find \textbf{directional biases} in four of five models that cannot be explained by simple floor or ceiling effects alone.
CD-$\uparrow$ should decrease, and CD-$\downarrow$ should increase, 
as source certainty rises: since texts that are already highly certain have little room to become more certain, and vice versa. 
However, because the \textit{low} and \textit{high} subsets are equally sized, these mechanical constraints alone do not account for the observed imbalance in directional changes. 
In SPICED, all models except \gengpt increase expressed certainty for \textbf{50--75\% of low-certainty source texts}, whereas they decrease expressed certainty for \textbf{20--50\% of high-certainty source texts}.
A similar asymmetry holds in MIMIC, where models increase the certainty of uncertain source texts in 25--30\% of examples but decrease the certainty of certain source texts in fewer than 16\%.
In both datasets, \genllama decreases certainty in certain texts at least as often as it increases certainty in uncertain texts.
Surprisingly, for already-confident texts, \genclaude and \gengemini still inflate certainty in a non-trivial fraction  (20\%) of SPICED examples.
These findings suggest that for some LMs, the bias towards ``overconfident'' outputs is not only associated with hedged inputs, but a more general tendency to strengthen the claims regardless of the source's starting point.

\emph{\textbf{Certainty Distortion is not a sampling artifact}}.
Our main results use top-p sampling (\texttt{top-p = 0.9} and \texttt{T = 1}), reflecting practical settings~\citep{Holtzman2020The}.
To distinguish model behavior from stochastic decoding effects,
we re-run experiments with greedy decoding (\texttt{T = 0}). Under greedy decoding, distortion is consistently higher, increasing by 9\%-20\% across models (\Cref{fig:decoding:breakdown:spiced}), with the largest increase for \genllama. 
Most additional distortions increase expressed certainty, indicating that greedy decoding amplifies the directional biases.
This contrasts with prior work showing that lower temperature reduces other forms of information distortion in content-constrained tasks~\citep{mohamed-etal-2025-llm,perez2025when}.

\subsection{The Broken Telephone Effect}
\label{ssec:sentence-level-iteration-certainty-distortion-rates}

The results so far focus on certainty distortion after a single LM transformation. We next ask whether repeatedly applying the same task compounds the observed distortion (the ``broken telephone'' effect), or if it remains the same. This has been investigated for information distortion~\citep{mohamed-etal-2025-llm,perez2025when} but not for certainty distortion. 
To examine this, we simulate a communication chain where each model rewrites its own previous output for up to five iterations, comparing each intermediate output to the original source text. 

\begin{figure}[tb]
    \centering
    \includegraphics[width=0.9\linewidth]{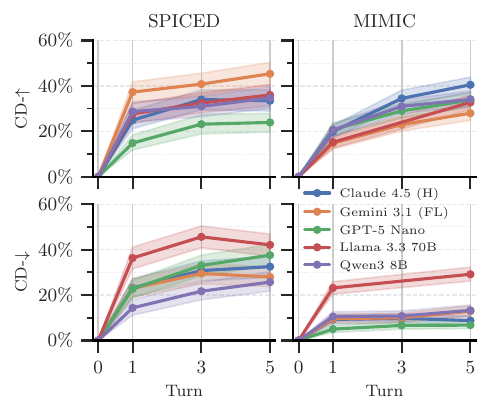}
    \caption{
    Certainty distortion across iterations (paraphrasing task) measured with respect to the certainty of the original text. 
    In SPICED (left), most distortion is observed within one LM call, with little relative change afterwards. 
    In MIMIC (right), \certincrease continues climbing after 5 LM calls, while \certdecrease remains flat, indicating directional compounding towards inflated certainty.
    }
    \label{fig:turn-effects:paraphrasing-tasks}
    \postspace
\end{figure}

As shown in \Cref{fig:turn-effects:paraphrasing-tasks}, in the scientific domain, most certainty distortion occurs after the first call: 
both \certincrease and \certdecrease rise sharply between turns 0 and 1, plateauing thereafter.
On the other hand, \textbf{in the medical domain, CD-$\uparrow$ continues to climb through turn 5}: \genclaude, for instance, goes from 20\% \certincrease after one iteration to 40\% after five. 
Meanwhile, CD-$\downarrow$ remains essentially flat for all models throughout, remaining under 15\% for four of five models, with \genllama as the exception. In other words, repeated paraphrasing in the medical domain compounds certainty distortion directionally, producing progressively more outputs with greater expressed certainty than the original source. 
{
One possible contributor is the greater density of uncertainty-bearing language in the medical sources: a descriptive lexicon-based analysis finds 4.84 epistemic markers per 100 words in MIMIC, compared with 1.54 in SPICED (Appendix \ref{apdx:broken-telephone-analysis}). This difference provides more opportunities for such cues to be progressively weakened, although it does not establish the mechanism behind the divergent trajectories.
}

{
Overall, the observed LM chain results raise the question of whether the stabilization and directional compounding patterns observed here would also emerge in human rewriting chains.
Prior work documents certainty distortion across stages of science communication but does not isolate fixed-task sequential rewriting chains~\citep{Brechman2009,Sumner2016,wuehrl-etal-2024-understanding}, while studies of human communication chains show that repeated retelling can preserve some dimensions while altering others~\citep{Breithaupt2018}, consistent with formal accounts of iterated learning~\citep{griffiths-kalish-language-evol-iterated-learning}.
Future work could address this gap by studying expressed certainty in fixed-task human rewriting chains, comparing instructions to preserve meaning alone with instructions to preserve both meaning and certainty.
}

\subsection{Distortion as a Function of LM Scale}
\label{ssec:model-scale}

So far, our analyses have focused on single models across 5 different model families. 
To assess whether distortion is an artifact of model scale or persists in larger, more capable models, we extend the analysis to bigger variants of \gengpt, \genclaude, and \gengemini. 
For cost reasons, we focus on the two tasks with the highest distortion rates in each domain: news rewriting in SPICED and paraphrasing in MIMIC. 
\Cref{tab:lm-scale-experiments} summarizes the results.

Overall, \textbf{\emph{even the largest evaluated LMs distort certainty}}, and the effects of model size vary across families and tasks. Among the largest variants, CD ranges from 46.1–69.5\% on SPICED and from 19.0–27.4\% on MIMIC. On news rewriting, CD drops from 74.8\% to 55.7\% across the Claude family and from 75.3\% to 69.5\% across Gemini variants, whereas the GPT family shows no overall improvement (44.3\% to 46.1\%). On medical paraphrasing, increasing model size produces no consistent improvement across families.\footnote{A BERT-based estimator recovers the GPT-variant ordering, reducing concerns about judge self-preference (Appendix~\ref{apdx:robustness-to-evaluator}).
}
Across all eight model variants, no model achieves distortion below 39.5\% or 11\% in SPICED and medical tasks, respectively. 
These results indicate that certainty distortion is a phenomenon that persists across LM scales.
\begin{table}[tb]
\small
\centering
\setlength{\tabcolsep}{3pt} 
\begin{tabular}{lcccc}
\toprule
& \multicolumn{2}{c}{\textbf{SPICED - News}} & \multicolumn{2}{c}{\textbf{MIMIC - Para}} \\
\cmidrule(lr){2-3} \cmidrule(lr){4-5}
\textbf{Model} & \textbf{\certaintydistortion} & \textbf{\certincrease}  & \textbf{\certaintydistortion} & \textbf{\certincrease} \\
\midrule
\texttt{gpt-5.4-nano}            & 44.3 & 23.2 & 25.8 & 20.8 \\
\texttt{gpt-5.4-mini}            & 39.5 & 15.1 & 11.1 & 9.8 \\
\texttt{gpt-5.4}                 & 46.1 & 15.1 & 19.0 & 16.2 \\
\addlinespace
\texttt{claude-haiku-4-5}        & 74.8 & 49.6 & 29.1 & 19.9 \\
\texttt{claude-sonnet-4-6}       & 66.8 & 40.1 & 27.0 & 11.4 \\
\texttt{claude-opus-4-7}         & 55.7 & 31.7 & 27.4 & 17.4 \\
\addlinespace
\texttt{gemini-3.1-flash-lt}     & 75.3 & 48.6 & 24.3 & 14.8 \\
\texttt{gemini-3.5-flash}        & 69.5 & 49.9 & 18.6 & 15.5 \\
\bottomrule
\end{tabular}
\caption{
    Effect of model scale on certainty distortion (\%) for news rewriting (SPICED) and paraphrasing (MIMIC). %
    Effects vary across model families and tasks, and no variant achieves below 11\% distortion.
}
\label{tab:lm-scale-experiments}
\postspace
\end{table}

\section{Beyond Single Sentence Tasks}
\label{sec:multi-sentence-experiments}

To test whether certainty distortion generalizes beyond sentence-level transformation, we extend our analysis to document-level tasks. 
We introduce a new dataset of 500 academic abstracts spanning five disciplines across social and life sciences, and use the full Findings section of 800 MIMIC-CXR reports, 
restricted to reports containing at least one uncertainty-bearing sentence to enrich the sample for certainty variation (Appendix \ref{apdx:sec:additional-dataset-details}).
We introduce a summarization task to simulate a literature review task, while extending the other tasks (paraphrasing and simplification) to the whole document.
Because outputs can rephrase, drop, or merge claims, we extend our LM-judge with an alignment step that locates output counterparts (if any) for finding-level claims in academic abstracts and sentences in medical reports before comparing certainty. 
{
We conduct a controlled sanity check of this evaluator using certainty-varied rewrites (Appendix~\ref{apdx:multi-sentence-experiments}) and leave full validation for future work.
}

\paragraph{Results.} 
Certainty distortion persists in document-level transformations, with patterns that echo our sentence-level findings: distortion is higher in the scientific domain than in the medical domain, and increases in expressed certainty generally occur more often than decreases. 
\Cref{tab:multi-sentence-results} shows that, across models, increases in expressed certainty occur at 3.1--11.3$\times$ the rate of decreases for academic abstracts. The asymmetry is weaker for medical reports, where ratios range from 0.9$\times$ to 4.6$\times$, with increases occurring at more than twice the rate of decreases for three of the five models, although overall distortion is lower (2--5\%). 
We view these results as initial evidence of document-level certainty distortion; independent validation of the alignment-based evaluator and investigation of the cross-domain gap are promising directions for future work.
\begin{table}[tb]
\centering
\small
\begin{tabular}{lcccc}
\toprule
& \multicolumn{2}{c}{\textbf{Academic Papers}} & \multicolumn{2}{c}{\textbf{MIMIC}} \\
\cmidrule(lr){2-3} \cmidrule(lr){4-5}
\textbf{Model} & \textbf{\certaintydistortion} & \textbf{\certincrease} & \textbf{\certaintydistortion} & \textbf{\certincrease} \\
\midrule
\gengpt         & 15.0 & 11.3$_{3.1}$       & 2.0 & 1.4$_{2.1}$ \\
\genclaude      & 22.8 & 21.0$_{11.3}$      & 3.4 & 2.8$_{4.6}$ \\
\gengemini      & 24.8 & 22.7$_{10.8}$      & 4.1 & 3.3$_{4.0}$ \\
\genqwen        & 18.4 & 15.8$_{6.3}$       & 5.0 & 2.7$_{1.1}$ \\
\genllama       & 15.8 & 12.3$_{3.6}$      & 4.5 & 2.1$_{0.9}$ \\
\bottomrule
\end{tabular}
\caption{
Certainty distortion (\%) on document-level scientific and medical tasks, micro-averaged over aligned claims and sentences from 500 abstracts and 800 reports, respectively. \certincrease denotes increased expressed certainty; subscripts report inflation-to-deflation ratios.
}
\label{tab:multi-sentence-results}
\postspace
\end{table}

\section{Prompt-Based Mitigation}
\label{sec:mitigation}

Now, we {
test whether standard, low-cost prompting interventions that require no access to model weights or training pipelines are sufficient to mitigate certainty shifts.}
Specifically, we test restricting the original prompts with additional constraints, including 
(1) \textit{preserve-semantics}, which instructs to keep meaning intact; 
(2) \textit{do-not-hallucinate}, which instructs against unsupported content;
and (3) \textit{preserve-certainty}, which explicitly instructs against shifting the certainty of the original text.
All three are appended to the base prompt; full text in Appendix \ref{apdx:prompts}.
We test the efficacy of these prompts on the scientific domain tasks because these exhibited highest CD.

\paragraph{Results.} 
Prompt interventions reduce distortion partially but do not eliminate it (\Cref{tab:mitigation:permodel-mean}).
Stacking all three constraints lowers \certaintydistortion from 50.6\% to 40.1\% on paraphrase and from 64.3\% to 41.9\% on news. 
However, we find that adding \textit{preserve certainty} brings the inflation-to-deflation ratio close to or below 1, while \textit{preserve semantics} and \textit{do not hallucinate} reduce both directions proportionally and leave the asymmetry intact. 

{
These results distinguish between correcting an aggregate directional bias and preserving certainty in individual rewrites. 
Although \textit{preserve certainty} approximately balances inflation and deflation, around 40\% of rewrites remain distorted, hence suggesting that zero-shot prompting rebalances errors rather than preventing them.
More effective mitigation may require instance-level feedback that explicitly compares rewrites with their source. Few-shot demonstrations could better specify certainty-preserving behavior, while verifier-guided generation or reranking~\citep{ichihara2025evaluation-2025-TMLR} could reject candidate rewrites that alter the source's epistemic stance. Training-based approaches could instead use supervised certainty-preserving rewrites or preference~\citep{rafailov2023direct} and reinforcement-learning objectives~\citep{Ouyang2022rlhf} that penalize such shifts.
}

\begin{table}[tb]
\small
\centering
\setlength{\tabcolsep}{3pt} 
\begin{tabular}{lcccc}
\toprule
 & \multicolumn{2}{c}{\textbf{Paraphrase}} & \multicolumn{2}{c}{\textbf{News}} \\
\cmidrule(lr){2-3} \cmidrule(lr){4-5}
\textbf{Intervention} & \textbf{\certaintydistortion} & \textbf{\certincrease} & \textbf{\certaintydistortion} & \textbf{\certincrease} \\
\midrule
original & 50.6 &  26.7$_{1.4}$ & 64.3 & 39.7$_{1.6}$  \\
\addlinespace
\quad + \textit{preserve semantics} & 47.4 & 25.8$_{1.4}$ & 51.0 & 33.1$_{1.9}$ \\
\quad \quad + \textit{do-not-hallucinate} & 42.8 & 24.6$_{1.4}$ & 47.3 & 30.7$_{1.9}$ \\
\quad \quad \quad + \textit{preserve-certainty} & 40.1 & 19.4$_{1.0}$ & 41.9 & 18.5$_{1.1}$ \\
\bottomrule
\end{tabular}
\caption{
Cumulative prompt interventions, averaged across models in SPICED. Subscripts on \certincrease report the inflation-to-deflation ratio.
Only \textit{preserve-certainty} comes close to neutralizing the directional bias.
}
\postspace
\label{tab:mitigation:permodel-mean}
\end{table}

\section{Conclusion}
\label{sec:conclusion}
LMs can distort human-generated text in ways that change the information conveyed \citep{mohamed-etal-2025-llm,abdulhai2026llmsdistortwrittenlanguage}. 
We focused on certainty distortion: shifts in expressed certainty that occur during rewriting intended to preserve meaning. 
Using an LM-based metric validated against human annotations, we characterized this phenomenon across five model families on scientific and medical domains. We find that certainty distortion affects a non-negligible fraction of LM outputs and is directionally asymmetric, with most models being more likely to inflate the certainty of the source than to attenuate it. This directional tendency compounds under repeated LM paraphrasing in the medical domain. Larger model variants and changes in decoding strategy do not consistently reduce distortion, while prompt-based interventions reduce but do not eliminate it. 
These results suggest that {
preserving the expressed certainty of aligned claims
} deserves attention as a first-class evaluation target alongside factuality and semantic fidelity.

\section*{Limitations}
\textbf{Cross-Domain and Cross-Lingual Generalization. } 
We evaluate certainty distortion only in English and within scientific and medical communication tasks. Although we observe distortion across both domains, the differences in magnitude suggest that domain characteristics, such as whether claims are general statements about the world versus patient-specific observations, play a meaningful role. Extending the analysis to additional domains (\eg legal, financial) and task types would help clarify how broadly these patterns generalize. Additionally, performing cross-lingual analysis would be a valuable direction, as hedging conventions and epistemic markers vary considerably across languages~\citep{Lauwereyns2002,Yagz2014,Shafqat2019,MurDueas2021}.

\textbf{Document-Level Evaluation Validity. } 
While our sentence-level evaluation is validated against human judgments, the document-level experiments rely on claim-alignment procedures that have not been comparably validated through human studies. To partially address this limitation, we assessed the medical evaluator using 160 controlled source--rewrite pairs that preserved, increased, or decreased certainty. Although this does not constitute an independent quantitative validation, its judgments generally followed the intended direction of the certainty shifts, providing preliminary evidence that it captures directional distortion (Appendix \ref{apdx:multi-sentence-experiments}).
We therefore view the document-level results as initial evidence and leave more rigorous human validation to future work. Techniques such as claim decomposition~\citep{wanner-etal-2024-closer,javaji-etal-2025-ai}, explicitly isolating individual propositions within a sentence before comparing certainty, could improve document-level evaluation in future studies. 

\textbf{Mechanisms of Certainty Distortion. }
Finally, although we observe lower one-turn distortion in the medical domain and greater directional compounding under repeated paraphrasing, our experiments do not identify the mechanism behind this cross-domain difference. 
{
Our analyses nonetheless constrain the space of explanations: distortion persists under greedy decoding (ruling out a sampling artifact), varies systematically by task and domain (implicating the communication setting), and exceeds what floor/ceiling effects alone predict.}
One hypothesis is that a model's internal belief about a claim, \ie whether it supports or opposes it, drives certainty inflation or deflation during rewriting. To test this, we conducted a belief elicitation experiment on the SPICED dataset: for each source claim, we collected model beliefs by normalizing certainty and stance into a single alignment score reflecting whether the model supported or opposed the claim, and compared this against the observed direction and magnitude of certainty distortion. We found no meaningful correlation (Kendall's $\tau_B \approx -0.1$), suggesting that prior belief alone does not explain the observed distortion patterns. 
An alternative explanation is that models internalize domain- and task-specific writing conventions during training. Science journalism, for instance, is known to conventionally favor confident, attention-grabbing language that strips hedges from original findings~\citep{Sumner2016,Dempster2022}, whereas radiology reports follow well-established conventions for expressing graded uncertainty~\citep{Lee2017,Shinagare2019}, and academic scientific writing treats hedging itself as a disciplinary norm~\citep{Hyland1998}. If models learn these register-specific patterns during pretraining, they may apply them during rewriting regardless of the source text's original certainty level. 
{
A third, non-exclusive hypothesis concerns post-training, as instruction tuning and preference optimization may favor responses that appear fluent, helpful, and decisive over  preserving epistemic uncertainty. 
Prior work comparing base, instruction-tuned, and RLHF models on question answering tasks finds that a preference for certainty strengtheners emerges after RLHF, while reward models and human preference data penalize expressions of uncertainty~\citep{zhou-etal-2024-relying}. Related tendencies of post-trained models toward sycophancy provide broader evidence that preference optimization can systematically alter how models communicate claims~\citep{perez-etal-2023-discovering,sharma2024towards-sycophancy}. However, these studies do not examine meaning-preserving rewriting and therefore do not establish post-training as the cause of the certainty distortion observed here.
}
Disentangling these potential mechanisms remains an important direction for future work.

\section*{Societal Impact}
{
Certainty distortion may have societal consequences even when rewritten text remains factually and semantically close to its source. In particular, certainty inflation can make tentative or weakly supported claims appear established, potentially amplifying misleading information and encouraging overreliance in high-stakes domains. Certainty deflation may obscure well-supported conclusions.
}

\section*{Acknowledgments}
We thank the members of MADLAB and UCINLP for their valuable feedback and discussions throughout this project. We are especially grateful to Yu Fei, Preethi Seshadri, Lukas Mayer, and ZhaoBin Li for their insightful comments. 
This work was supported by National Science Foundation under NSF CAREER award IIS-2046873 and award NSF 2505006, by the Hasso Plattner Institute (HPI) Research Center in Machine Learning and Data Science at UCI, and by funding support from Google and SAP.
The views and conclusions expressed in this paper are those of the authors and should not be interpreted as representing the official policies, views, or endorsements of the funding agencies or organizations.

\bibliography{custom}

\appendix
\section{Automatic Metric Validation}
\label{apdx:automatic-metric-validation}

Automatic metric validation proceeded in two stages. 
First, we began with a validation of different certainty estimators (\ie LLM-judge approach, and fine-tuned BERT models~\citep{pei-jurgens-2021-measuring,wuehrl-etal-2024-understanding}) in a set of synthetically generated datasets, controlled for levels of certainty distortion. Once we were comfortable with the proposed approaches, we conducted the user study to validate our findings.  The remainder of this section is organized as follows:
\begin{itemize}
    \item Appendix \ref{apdx:qualitative-analysis} provides concrete examples where BERT~\citep{wuehrl-etal-2024-understanding} and our proposed LM-as-a-judge approach disagree.
    \item Appendix \ref{apdx:controlled-eval} describes the controlled experiments conducted earlier on in this project that led to the final proposed LM-as-a-judge metric, using \evalgpt.
\end{itemize}

\subsection{Qualitative Analysis}
\label{apdx:qualitative-analysis}

\Cref{tab:qualitative-pairwise-examples:pei-jurgens-lm-judge,tab:qualitative-pairwise-examples:wuerl-et-al-lm-judge} represent examples where the proposed LM-as-a-judge metric and two existing BERT-based metrics disagree. 
In addition to listing the original (or source text) and an LM-generated text, we also report the predicted scores by the two evaluators. 
\begin{table*}[tb]
\small
\centering
\resizebox{\linewidth}{!}{
\begin{tabular}{p{0.40\linewidth} p{0.40\linewidth} p{0.07\linewidth} p{0.07\linewidth} }
\toprule
\textbf{Original} & \textbf{LM-generated} & \textbf{BERT} & \textbf{LLM} \\
\midrule
'In particular, we find that a strong social media presence \textbf{may be} more important than being under contract with a top agency, or than the aesthetic standards sought after by the industry.' & '\textbf{Notably}, having a strong presence on social media \textbf{can matter} more than being signed with a leading agency or meeting the fashion industry’s preferred look.' & -1.0 & 1.5 \\ 
\rowcolor{gray!10} 'To the best of our knowledge, this is the first convincing demonstration of a professional group showing higher accuracy on face matching tasks.' & 'A notable breakthrough has been achieved, with a particular profession exhibiting unparalleled precision in identifying and matching facial features.' & -0.97 & 2.0 \\ 
\cmidrule{1-4}
'Yield increase is therefore shown to be more effective in reducing land use when combined with land-use regulation and consumption changes.' & 'The results suggest that pairing higher yields with policies that regulate land use and shift consumption patterns more effectively lowers land demand.' & 0.62 & -1.0 \\ 
\rowcolor{gray!10} 'As exercise capacity improves, progression towards longer sessions of at least 20 min duration on most days of the week is recommended.' & 
'It is suggested that as physical fitness increases, individuals should aim to engage in extended periods of exercise, ideally lasting 20 minutes or more, on a majority of days throughout the week.' & 0.62 & -1.0 \\
\cmidrule{1-4}
'Across three studies, well-educated graduates who entered the workforce during economic downturns were more satisfied with their current jobs than those who entered during more prosperous economic times.' & 
'Three studies revealed that highly educated graduates entering the job market during economic recessions reported greater job satisfaction compared to those who joined during more prosperous periods.' & 1.0 & 0.0 \\ 
\rowcolor{gray!10}  'Teens starting school at 8:30 AM or later are the only group with an average time in bed permitting 8 hours of sleep, the minimum recommended by expert consensus for health and well-being.' &  'Experts agree that teenagers need at least eight hours of sleep for optimal health, and only those whose school day begins at 8:30 AM or later typically get enough rest.' & 1.0 & 0.0 \\ 
\cmidrule{1-4}
The towns and villages that constituted research sites for our study are \textbf{likely} similar to hundreds or even thousands of sites throughout Asia, where humans live alongside synanthropic macaques. & Our study's research sites, comprising towns and villages, \textbf{likely} resemble hundreds or thousands of similar locations across Asia where human populations coexist with synanthropic macaques. & 0.89 & 0.5 \\
\rowcolor{gray!10} Study anticoagulant was discontinued because of bleeding in 242 patients (4.7\%) in the otamixaban group and in 95 patients (1.7\%) in the UFH-pluseptifibatide group (P < .001). & The study found that significantly more patients in the otamixaban group had to stop treatment due to bleeding compared to those in the UFH plus eptifibatide group. & 0.99 & -0.5 \\
\bottomrule
\end{tabular}}
\caption{Qualitative analysis of edge cases where the reference-less BERT-based model proposed by~\citet{pei-jurgens-2021-measuring} disagrees with the pairwise-ranking LM-judge approach proposed in the paper. The first four examples were selected from extreme cases, and the remaining four were randomly picked.}
\label{tab:qualitative-pairwise-examples:pei-jurgens-lm-judge}
\end{table*}

\begin{table*}[tb]
\small
\centering

\resizebox{\linewidth}{!}{
\begin{tabular}{p{0.40\linewidth} p{0.40\linewidth} p{0.07\linewidth} p{0.07\linewidth} }
\toprule
\textbf{Original} & \textbf{LM-generated} & \textbf{BERT} & \textbf{LLM} \\
\midrule
'Underlying consolidation would be difficult to exclude.' & 'It is hard to rule out lung consolidation that is not seen on imaging.' & 2.0 & -0.5 \\ 
\rowcolor{gray!10} 'The cardiac silhouette \textbf{appears} top normal in size.' & 'The heart is a normal size.' & 0.0 & 2.0 \\ 
\cmidrule{1-4}
'Please note that pericardial fluid would be difficult to exclude on this study.' & 'It is difficult to determine whether or not there is fluid around the heart based on this scan.' & -1.0 & 2.0 \\ 
\rowcolor{gray!10} 'Increased interstitial markings which \textbf{could} represent a component of interstitial edema or chronic lung disease.' & 
'The radiographic image shows enhanced interstitial patterns, \textbf{suggestive} of either interstitial fluid accumulation or a chronic pulmonary condition.' & 2.0 & -0.5 \\
\cmidrule{1-4}
'Some focal opacity in the right upper zone laterally is noted -- \textbf{if real}, this is new, but it \textbf{could} represent artifact due to the presence of the subcutaneous emphysema.' & 
'A small area of abnormal density is visible in the outer part of the right upper lung zone, although it is \textbf{unclear} whether this is a genuine finding or an artifact caused by air trapped under the skin.' & 2.0 & -0.5 \\ 
\rowcolor{gray!10}  'Airspace consolidation is noted within the right lower lobe \textbf{concerning} for pneumonia.' &  'The chest X-ray shows a cloudy area in the lower right lung, \textbf{indicating a strong likelihood} of pneumonia.' & 0.0 & 2.0 \\ 
\bottomrule
\end{tabular}}
\caption{Qualitative analysis of edge cases where the reference-less fine-tuned BERT-based model proposed by~\citet{wuehrl-etal-2024-understanding} disagrees with the pairwise-ranking LM-judge approach proposed in this paper. The first three examples are drawn from the MIMIC-CXR~\citep{Johnson2019-MIMIC-CXR} paraphrasing task, and the remaining three are drawn from the simplification task.}
\label{tab:qualitative-pairwise-examples:wuerl-et-al-lm-judge}
\end{table*}

\subsection{Controlled Evaluation}
\label{apdx:controlled-eval}

With the goal of better understanding how to detect differences in certainty, we conduct a control experiment where for each scientific finding in SPICED, we create three controlled subsets: 
\begin{itemize}
    \item \spicedcontrolsame: pairs of texts where the confidence in the modified text $x_M$ is equivalent to the original text $x_o$, \ie $\delta(x_o, x_M) \approx 0$;
    \item \spicedcontrolless: pairs of texts where the confidence in the modified text $x_M$ is explicitly lower than the original text $x_o$, \ie $\delta(x_o, x_M) < 0$;
    \item \spicedcontrolmore: pairs of texts where the confidence in the modified text $x_M$ is explicitly higher than the original text $x_o$, \ie $\delta(x_o, x_M) > 0$.
\end{itemize}

\subsubsection{Synthetic Control Dataset Construction} 
\label{appdx:validation-uncertainty-estimator:dataset-creation}

%
This section describes the creation process for the controlled datasets.
We restrict this analysis to the examples in SPICED~\citep{wright-etal-2022-SPICED} dataset, whose final information matching score is less than 3. Note that this represents the complement of the subset used in the main paper, which concerned the subset whose score was $\geq 4$. 
Additionally, we filter out incomplete sentences (\eg ``As illustrated in the scatter plot ( Fig.'') using spaCy's constituency parser~\citep{spacy}. 
Finally, we randomly sample 400 scientific findings, which we use to create the synthetic datasets.

To obtain the three variants, we prompt \gptbig and \geminibig to perform targeted minimal edits s.t. only the epistemic language associated with the main reported finding is affected. Both models are called with \texttt{T=0.5}, providing a good trade-off between diversity and prompt-adherence.\footnote{Experiments were conducted throughout March 2026.} All implementation details will be shared in a public repository.
%
%
%
%

The resulting controlled datasets provide pairwise rankings of epistemic certainty rather than absolute scores. This choice is motivated by the subjectivity of certainty judgments, which often leads to low inter-annotator agreement~\citep{pei-jurgens-2021-measuring,wuehrl-etal-2024-understanding}. Additionally, prior work has shown that relative comparisons yield more reliable judgments than absolute scoring across tasks such as machine translation~\citep{Callison-Burch-WMT-RelativeRanking,moosa2024mtranker}, summarization~\citep{zopf-2018-estimating,luo-etal-2022-prefscore}, and other generative evaluation settings~\citep{chatbot-arena-2023}. 
With these controlled datasets we can evaluate whether a certainty scoring function can detect \textit{directional changes in certainty}, assigning higher scores to the more certain variant and lower scores to the less certain variant within each pair, while being able to distinguish true paraphrases that preserve the certainty.
\subsubsection{Dataset Statistics and Validation}
\label{appdx:validation-uncertainty-estimator:dataset-validation}

This section validates that the generated variants 
    (1) preserve the semantics of the original sentences and 
    (2) modify the epistemic language such that the text is perceived as more certain in \spicedcontrolmore and less certain in \spicedcontrolless.

\paragraph{Length Analysis.} 
We begin by measuring differences in average number of sentences between the original scientific finding and the modified variant. 
Using \texttt{nltk.sent\_tokenize} to partition the text, we find all outputs to have matching length, suggesting that modifications remain at the sentence-level.

\begin{figure}[tb]
    \centering
    \includegraphics[width=1\linewidth]{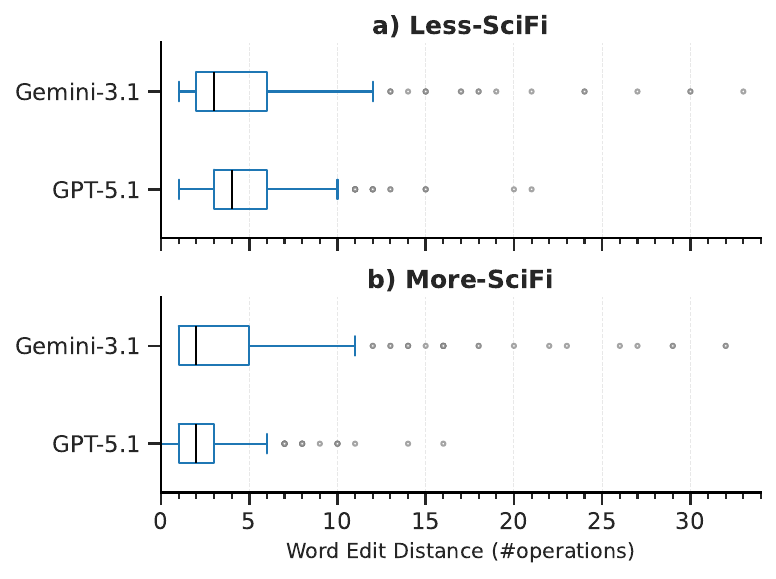}
    \caption{Word Edit Distance (WED) for the two SPICED control datasets. We discriminate the number of operations by model. We observe that the median number of modifications ranges between 3 and 4 in \spicedcontrolless and 2 for the \spicedcontrolmore.}
    \label{fig:spiced-control-word-analysis-edit-distance}
\end{figure}

\paragraph{Word Analysis.} 
The number of sentences provides only a coarse signal about the extent of the modifications. To better capture how much the text changes, we also evaluate the Word Edit Distance (WED), which measures the minimum number of word-level operations--additions, removals, or replacements--required to convert one sentence into another. In our datasets, models are instructed to produce minimally edited versions of the original sentence. Therefore, we expect WED values to be low, meaning that only a few operations are needed to transform the original sentence into the LM-generated variants.
WED is computed on lowercase, non-punctuated text using \texttt{rapidfuzz.distance.Levenshtein}. 
\Cref{fig:spiced-control-word-analysis-edit-distance} breaks down the WED statistics by dataset and models.
Considering \spicedcontrolless, we observe that the WED generally ranges from 2 to 6 operations. These edits include replacing assertive language such as ``is'', ``has'', ``showed'', ``significantly'', ``improved'', and ``strongly'' with modal verbs like ``may'', ``might'', and ``could'', as well as other less certain expressions such as ``appeared'', ``appears'', ``suggest'', and ``potentially''. 
We also manually examine examples with WED > 6 and find that higher WED values are typically associated with replacing multiple occurrences of epistemic language within the same sentence or replacing a single word with a multi-word phrase (\eg ``point to'' is replaced with ``seem to point to'', ``fits'' is replaced with ``appears to be consistent with'').
When considering \spicedcontrolmore, we observe slightly lower WED values, indicating that fewer modifications are required compared to \spicedcontrolless. In many cases, it suffices to add adverbs such as ``definitely'', ``clearly'', or ``strongly'' to increase the expressed certainty. We also observe that modal verbs (\eg ``may'', ``can'', might'', should'') and hedges (\eg ``possible'', ``believe'', ``appears'', ``suggests'') are often replaced with more assertive verbs such as ``confirms'', ``are'', ``demonstrate'', and ``establish''. As before, we find higher WED values to be associated with replacing a word or short phrase with a longer expression (\eg ``strong majorities'' replaced with ``It is a demonstrated fact that'').

\begin{figure}[tb]
    \centering
    \includegraphics[width=1\linewidth]{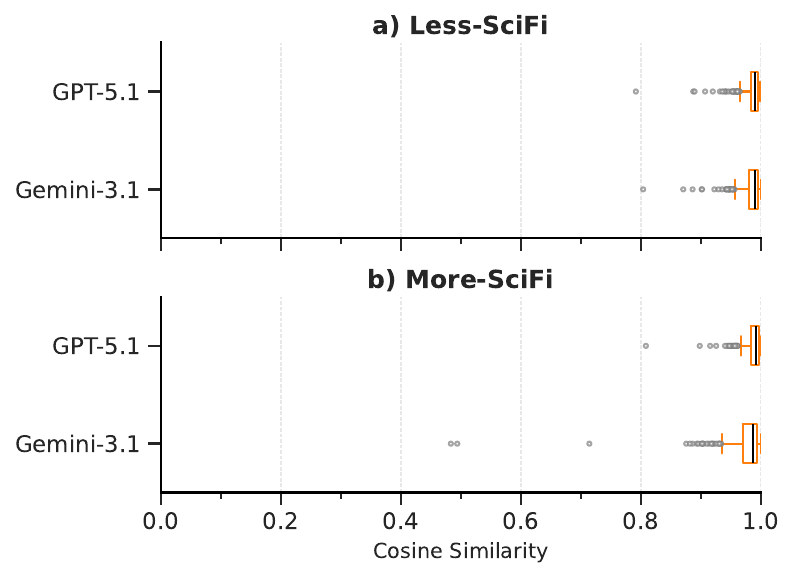}
    \caption{Semantic similarity analysis for \spicedcontrolless and \spicedcontrolmore. The median cosine similarity is close to 1 across models and dataset variants, indicating high average similarity between the source and transformed texts.}
    \label{fig:spiced-control-semant-analysis-cosine-similarity}
\end{figure}

\paragraph{Semantic Analysis.}
The word-level analysis indicates that LM-generated sentences are minimally edited relative to the originals, but it provides limited insight into whether their semantics are preserved. To quantify semantic preservation, we also measure pairwise cosine similarity in the control datasets using \texttt{MongoDB/mdbr-leaf-mt}~\citep{mdbr_leaf}, a compact, fast text embedding model that is competitive with OpenAI's text-embedding-3-small. Intuitively, if the generated outputs retain the meaning of the original SPICED sentence, the cosine similarity is expected to be close to 1.
\Cref{fig:spiced-control-semant-analysis-cosine-similarity} shows high cosine similarity across both control datasets.
Upon manually inspecting examples with lower cosine similarity (<0.9; n = 6), we find that these cases are typically either shorter sentences, which are more sensitive to small word changes (\eg ``It thus does not fit our definition.'' versus ``It thus may not fully fit our definition.'') or examples where the register differs, such as a shift from informal spoken language to formal written style. 
%
%
Interestingly, we observe similar patterns for the examples in \spicedcontrolmore with lower cosine similarity (<0.9; n = 10). We also find that the lower similarity scores in 4 out of 10 examples are associated with incomplete sentences. This can be attributed to our prompting design, which generates both certainty variants in a single model call while restricting the number of output tokens.
In the analysis that follows, we remove examples with lower cosine similarity from our control datasets, as they seem to correspond to incomplete or otherwise sensitive generations, totaling n=790 and n=786 examples for \spicedcontrolless and \spicedcontrolmore, respectively.

\paragraph{Manual Validation.}
To complement the automated evaluation, three authors of this paper manually annotated 150 paired examples, equally balanced across the three variants. We observe substantial agreement among annotators, as indicated by a Fleiss' kappa of $\kappa = 0.68$~\citep{Fleiss1971}. 
A more fine-grained analysis across subsets shows that agreement is higher for \spicedcontrolless and \spicedcontrolmore ($\kappa \approx 0.74$), but substantially lower for \spicedcontrolsame ($\kappa \approx 0.22$).
This lower agreement is expected, as examples in \spicedcontrolsame exhibit minimal or no differences in certainty, making the annotation task inherently more ambiguous.

\subsubsection{Evaluation of Certainty Estimators} 
\label{appdx:validation-uncertainty-estimator:eval-certainty-estimator}

With these controlled datasets, we can now compare how well different measures of uncertainty capture the directional biases in the control data. 
Specifically, we evaluate previously proposed BERT-based fine-tuned models~\citep{pei-jurgens-2021-measuring,wuehrl-etal-2024-understanding} as well as various prompting approaches.\footnote{
We evaluate existing estimators, leaving the development of specialized pairwise discriminative models to future work.}

\paragraph{A.2.3.1. Aspect-Level Certainty Estimates}
We start by examining the outputs of BERT-based models fine-tuned to estimate certainty in the context of scientific communication.
Through several rounds of annotation, ~\citet{pei-jurgens-2021-measuring} collect human annotations of certainty in scientific findings across various fields, achieving an average Krippendorff's $\alpha$ of 0.57. The annotations cover 6 certainty aspects:  Condition, Extent, Framing, Numerical, Probability, Suggestion. 
Using a subset of this (imbalanced) annotated dataset, the authors fine-tune a SciBERT model~\citep{beltagy-etal-2019-scibert} to predict a categorical distribution over three classes--Uncertain, NotPresent, Certain--for each aspect. 
The proposed model outperforms four other baselines\footnote{
These include bag-of-words and hedge-frequency baselines. 
\citet{pei-jurgens-2021-measuring} find that these baselines underperform model-based estimators and that hedges capture only part of expressed certainty. We therefore restrict our analysis to their stronger model-based estimator.
} across different aspects,
observing lower F1-scores for less common aspects such as Condition or Suggestion.  For that reason, we only consider the model predictions for four aspects: Extent, Framing, Number, and Probability.

In the main paper, we hypothesize the existence of an \textit{oracle certainty scoring function} that maps any text to an absolute certainty score, $s: \mathcal{X} \mapsto [0, 1]$. However, this differs from the fine-tuned SciBERT model, which outputs a probabilistic distribution over three classes for each aspect (mathematically, $f: \mathcal{V} \mapsto [0, 1]^3$ where $\sum_if[i] = 1$, where $f$ is an aspect-specific classifier) and therefore requires an additional transformation to produce a scalar score. 
%
%
\paragraph{Methodology.} 
One simple approach is to consider the class-wise difference. Given two sentence-level inputs $u$ and $v$, we compute $\Delta C(u, v; i) = f(v)[i] - f(u)[i]$, where $[i]$ denotes the $i$-th class.
However, by focusing on a single class (\eg Certain), this metric fails to account for the probability mass assigned to the other classes. To illustrate why this matters, consider the example where $f(u) = [0.4, 0.5, 0.1]$ and $f(v) = [0.7, 0.2, 0.1]$. If we only examine the difference in the Certain class (Class 2), we overlook the substantial shift in probability mass for the Uncertain class (Class 0), which may indicate that $u$ expresses less certainty than $v$.
An alternative approach is to aggregate the Certain and Uncertain scores into a continuous certainty measure that (1) is high when the Certain class dominates and (2) is low when the Uncertain class dominates. One possible formulation is $C(u) = f(u)[2] - f(u)[0]$, where $C(u) \in [-1, 1]$.

\begin{figure}[tb]
    \centering
    \begin{subfigure}{\linewidth}
        \centering
        \includegraphics[width=\linewidth]{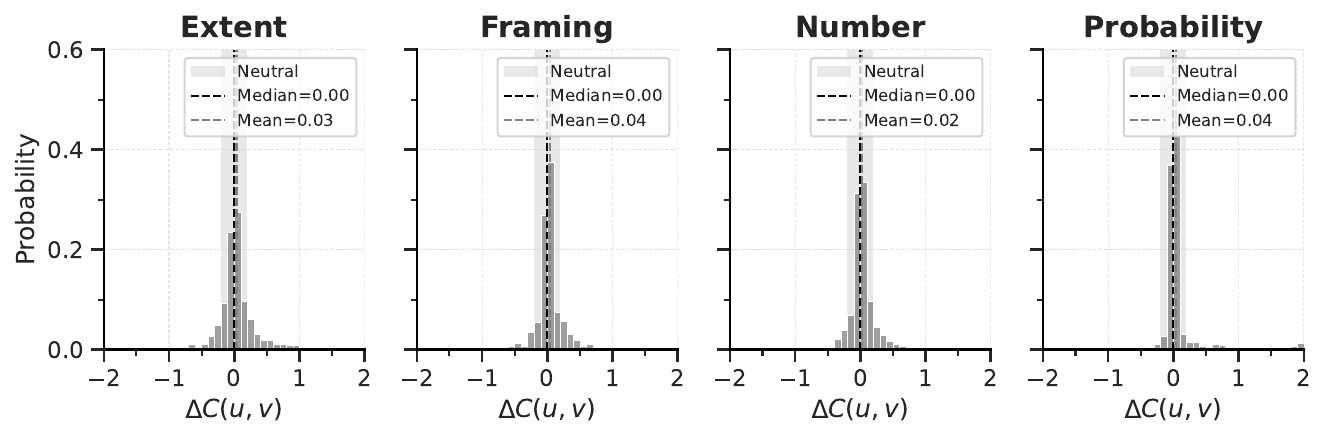}
        \caption{\spicedcontrolsame}
    \end{subfigure}
    \begin{subfigure}{\linewidth}
        \centering
        \includegraphics[width=\linewidth]{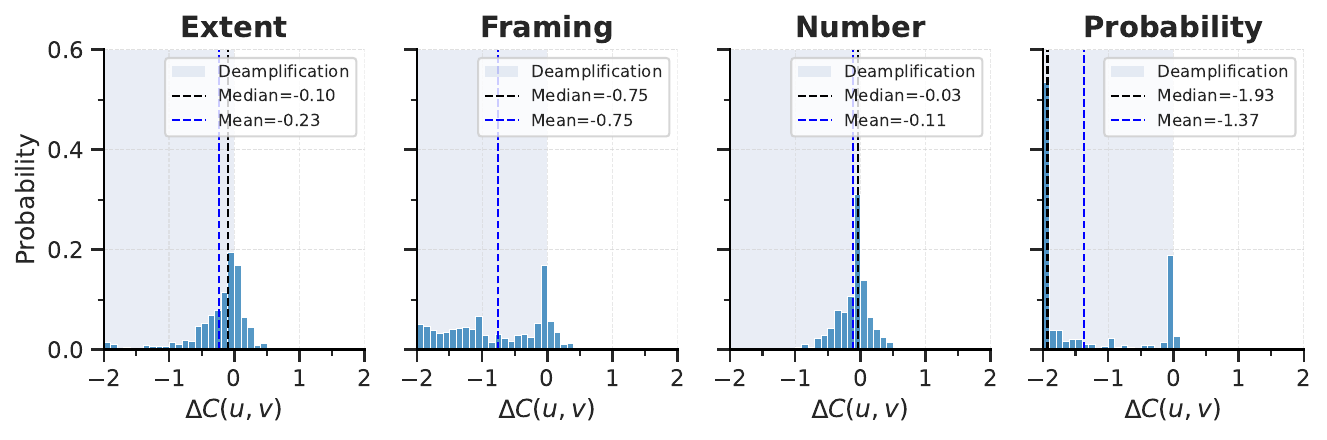}
        \caption{\spicedcontrolless}
    \end{subfigure}
    \begin{subfigure}{\linewidth}
        \centering
        \includegraphics[width=\linewidth]{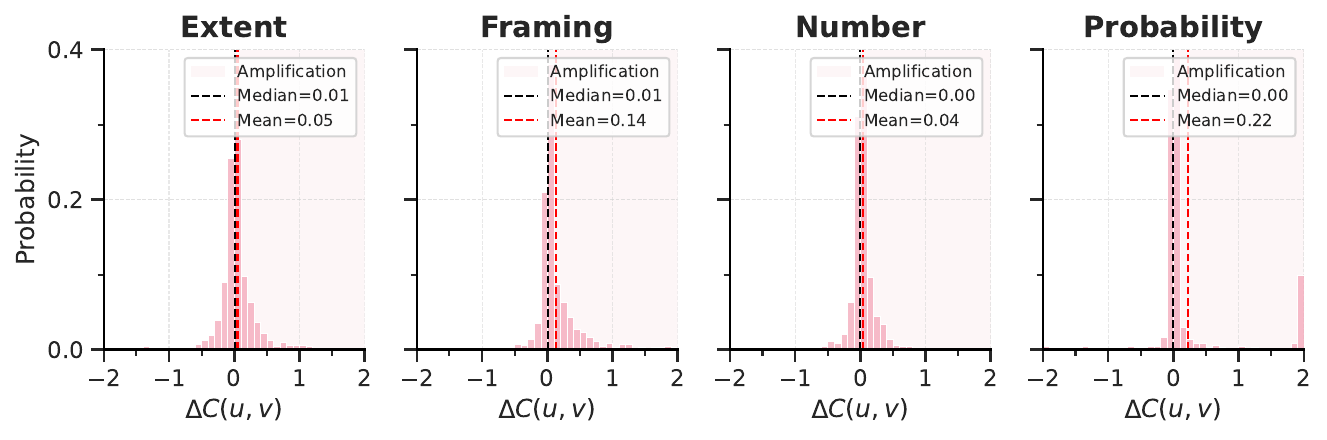}
        \caption{\spicedcontrolmore}
    \end{subfigure}
    
    \caption{Distribution of distortion scores in sentence-level control datasets. Certainty scores are computed using the aspect-level certainty classifiers~\citep{pei-jurgens-2021-measuring}. Given the original scientific finding $u$ and its modified version $v$, we report $\Delta C(u, v) = C(v) - C(u)$. We find that scores from Framing and Probability classifiers can be useful to detect decrease in certainty (with medians ranging between -0.75 and -1.93) but that they are not able to detect increase in certainty (with medians close to 0).}
    \label{fig:latex-figures-spiced-control--pei-jurgen-2021-aggr-score}
\end{figure}

\paragraph{Results.}
Having defined a certainty score for each sentence-level input, we can estimate the certainty distortion for each dataset by averaging the example-level differences. Because of the way we constructed the controlled datasets, we expect a reliable certainty scoring function to correctly reflect the lower certainty in \spicedcontrolless and higher certainty in \spicedcontrolmore. 
\Cref{fig:latex-figures-spiced-control--pei-jurgen-2021-aggr-score} shows the distortion gap distribution for both datasets across different aspects ($\Delta C(u, v; f) = C(v) - C(u)$). While the distortion gap is noticeably left-skewed for Framing and Probability in the \spicedcontrolless (with median values of -0.75 and -1.93), only negligible differences are observed for Framing and Probability in \spicedcontrolmore (with median values around 0.0 but mean values superior to 0.14). 
%
%

\begin{figure}[tb]
    \centering
    \includegraphics[width=\linewidth]{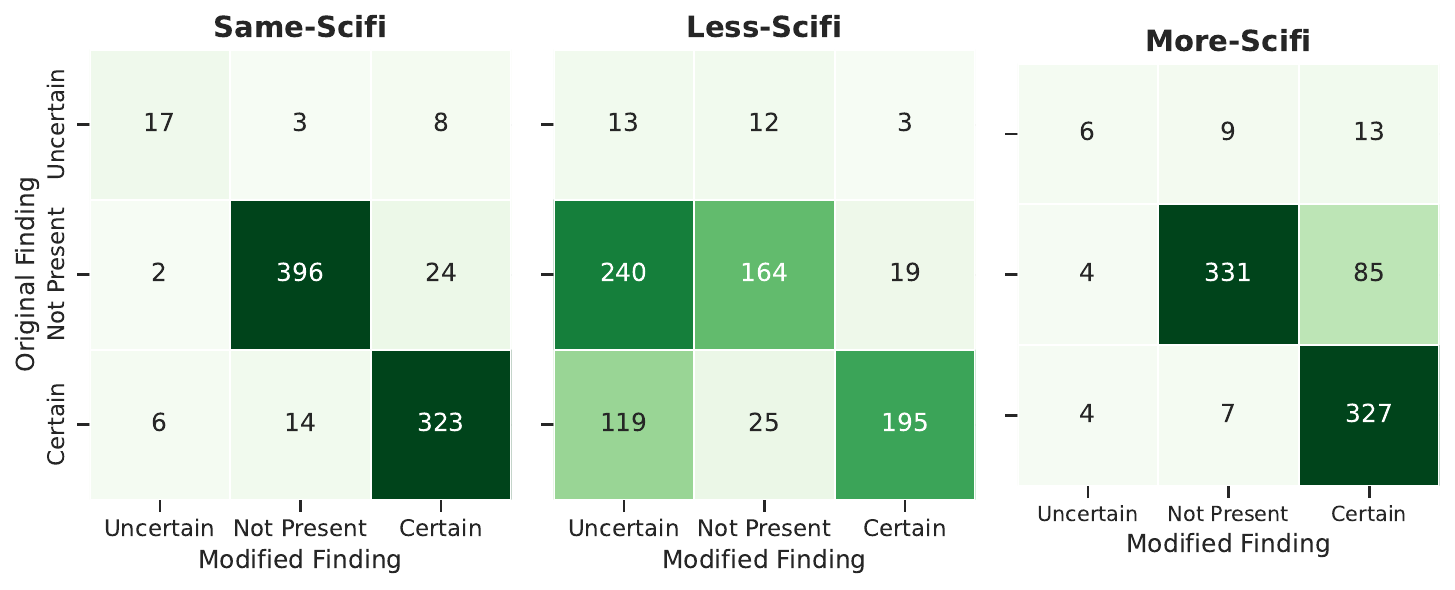}    
    \caption{Class predictions using the Framing certainty classifier proposed in \citet{pei-jurgens-2021-measuring}. Modified predictions are collected by considering the \textit{maximum a posteriori}. Unlike the \spicedcontrolless where model classes become less certain (lower triangular), in \spicedcontrolmore there are no significant class shifts, which suggests the classifier lacks granularity to detect the increase in certainty. }
    \label{fig:latex-figures-spiced-control--pei-jurgen-2021-Framing-certain-only-conf-matrices}
\end{figure}

\paragraph{Analysis.} To ensure that our findings are not specific to the scoring function $C(u) = f(u)[2] - f(u)[0]$, we repeat the analysis using a simpler certainty-specific function, $C(u) = f(u)[2]$. However, this metric also fails to accurately capture the increase in certainty in \spicedcontrolmore. 
To better understand this behavior, we examine the class boundaries of the Framing certainty classifier. We observe that, unlike in the \spicedcontrolless, there is little difference in the class distribution for \spicedcontrolmore (see \Cref{fig:latex-figures-spiced-control--pei-jurgen-2021-Framing-certain-only-conf-matrices}). 
This lack of sensitivity may stem from the nature of modifications in \spicedcontrolmore. As discussed earlier, these examples are often constructed by adding adverbs and adjectives that emphasize certainty (\eg strongly, definitely, significantly). We hypothesize that for sentences already receiving high certainty scores ($f[u] > 0.2$) the addition of such terms does not further increase the predicted certainty. 
%
For instance, the original finding ``These findings \textbf{suggested} the importance of liver pathogenesis in SMA.'' is assigned a certainty score of $f(u)[2] = 0.997$, whereas the modified version ``These findings \textbf{strongly support} the importance of liver pathogenesis in SMA.'' is attributed a score of $f(v)[2] = 0.999$. Another example is ``These examples \textbf{demonstrate} that some introns significantly boost expression, even in the absence of prior promoter activity.'' ($f(u)[2] = 0.98862$) versus the modified version ``These examples \textbf{definitively prove} that some introns significantly boost expression, even in the absence of prior promoter activity.'' ($f(v)[2] = 0.99762$).
This saturation effect may arise from limitations in the training data, which may not adequately capture the impact of these linguistic cues, and suggests that the classifier lacks the granularity needed to detect increases in certainty.

%
%
%
\paragraph{A.2.3.2. Global Certainty Estimates.} 
In a similar vein, ~\citet{wuehrl-etal-2024-understanding} fine-tunes a SciBERT model on scientific findings annotated with a single certainty score. The annotations are collected in a 4-point Likert Scale ranging from (1) Uncertain to (4) Certain. Inter-annotator agreement is moderate (average Spearman Correlation Coefficient is 0.44), suggesting task difficulty and subjectivity of the task.
Noticing the lower class prevalence of the Uncertain class, we combine the lower certainty classes into a single Uncertain class. The resulting class prevalence in training (n=1330) is 21:32:47, whereas in the test set (n=334 examples) is 25:33:42 for (0) Uncertain, (1) Somewhat Certain, and (2) Certain classes, respectively. 
%
%
%
\paragraph{Methodology.}
Since the original models are not publicly available, we fine-tune \texttt{allenai/scibert\_scivocab\_uncased} for certainty prediction. In addition to using the default hyperparameter configuration, we perform Bayesian hyperparameter optimization over the modeling objective (classification vs ordinal regression), learning rate, warm-up steps, and number of frozen layers, training a total of 80 models. Given the ordinal structure of the labels, we select the final model by \textbf{maximizing the quadratic weighted kappa (QWK)} on the test set. 
%
%
QWK measures the agreement between two annotators while accounting for the ordinal relationship between labels, penalizing larger disagreements more heavily than smaller ones. It takes values in $(-1, 1)$, where higher values indicate stronger agreement, a value of 0 corresponds to chance agreement, and negative values indicate systematic disagreement. 
\Cref{tab:wurl-et-al-best-bert-metric-classif-vs-ordinal} shows the final performance metrics on the test set for the classification and ordinal models with highest QWK across all runs.  
\begin{table}[t]
\centering
\small
\setlength{\tabcolsep}{4pt}
\begin{tabular}{l cc}
\toprule
\textbf{Metric} & \textbf{Classification Bert} & \textbf{Ordinal Bert} \\
\midrule
QWK & 0.551 & 0.500 \\
Loss & 0.939 & 0.957 \\
Accuracy & 0.608 & 0.580 \\
F1 (macro) & 0.601 & 0.560 \\
Precision (macro) & 0.603 & 0.570 \\
Recall (macro) & 0.599 & 0.560 \\

\cmidrule(lr){1-3}
\rowcolor{sectionbg}
\multicolumn{3}{l}{\textit{Per-class F1}} \\
\quad Class 0 & 0.621 & 0.540 \\
\quad Class 1 & 0.493 & 0.460 \\
\quad Class 2 & 0.688 & 0.680 \\

\cmidrule(lr){1-3}
\rowcolor{sectionbg}
\multicolumn{3}{l}{\textit{Per-class Precision}} \\
\quad Class 0 & 0.641 & 0.580 \\
\quad Class 1 & 0.500 & 0.470 \\
\quad Class 2 & 0.669 & 0.650 \\

\cmidrule(lr){1-3}
\rowcolor{sectionbg}
\multicolumn{3}{l}{\textit{Per-class Recall}} \\
\quad Class 0 & 0.602 & 0.510 \\
\quad Class 1 & 0.487 & 0.460 \\
\quad Class 2 & 0.707 & 0.710 \\
\bottomrule
\end{tabular}
\caption{Comparison of performance between the classification model and the ordinal-head model. The reported macro F1 results are on par with the values reported in the original paper (mF1 = 0.53 in Table 3 in ~\citet{wuehrl-etal-2024-understanding}).}
\label{tab:wurl-et-al-best-bert-metric-classif-vs-ordinal}
\end{table}

\begin{figure}[tb]
    \centering
    \begin{subfigure}{\linewidth}
        \centering
        \includegraphics[width=\linewidth]{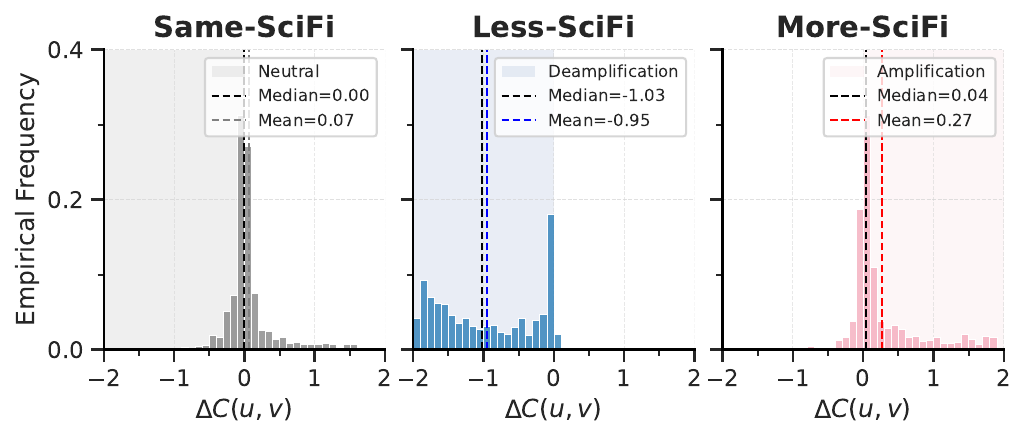}
        \caption{Classification SciBERT}
        \label{fig:latex-figures-spiced-control--LessSciFi-MoreSciFi--wurl-et-al-2024-classif-aggr-score}
    \end{subfigure}
    \vspace{1em}
    \begin{subfigure}{\linewidth}
        \centering
        \includegraphics[width=\linewidth]{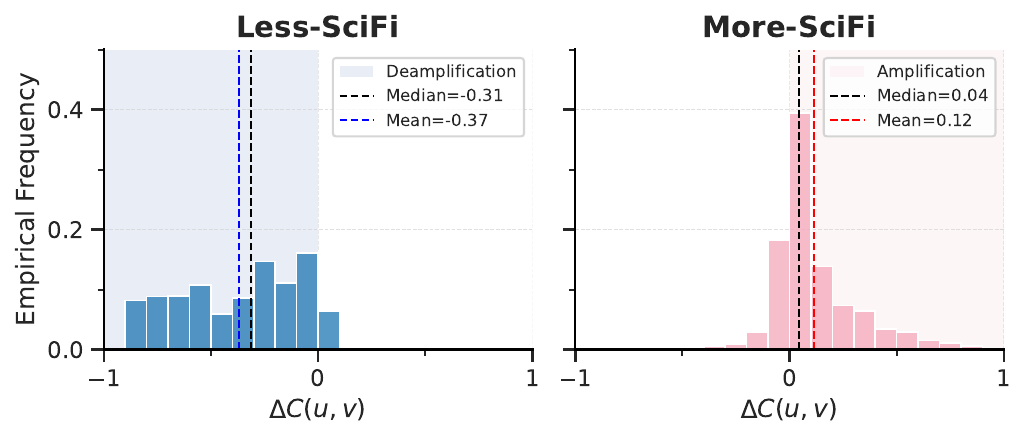}
        \caption{Ordinal Regression SciBERT}
        \label{fig:latex-figures-spiced-control--LessSciFi-MoreSciFi--wurl-et-al-2024-ordinal-aggr-score}
    \end{subfigure}
    \caption{Distribution of certainty distortion scores in the \spicedcontrolless and \spicedcontrolmore. Certainty scores are computed using SciBERT models trained to predict a single absolute certainty score~\citet{wuehrl-etal-2024-understanding}. The score for a single sentence $u$ is computed as $C(u) = f(u)[2] - f(u)[0]$, where the two terms correspond to the likelihood of the certain and uncertain classes for sentence $u$ according to the classifier $f$.}
    \label{figures/spiced-control--LessSciFi-MoreSciFi--wurl-et-al-2024-models-aggr-score}
\end{figure}

\paragraph{Results.} 
We now assess how well each of the two models can detect the directional certainty shifts in the controlled SPICED datasets. Because the classification model predicts a 3-class distribution, we compute the certainty score for a single sentence $u$ as the class-wise probability difference in the extreme classes, \ie $C(u) = f(u)[2] - f(u)[0]$. 
\Cref{fig:latex-figures-spiced-control--LessSciFi-MoreSciFi--wurl-et-al-2024-classif-aggr-score} shows the certainty distortion observed in both datasets. While this method allows us to correctly detect the decrease in certainty in \spicedcontrolless (median=-1.03), it is not a reliable measure of confidence amplification in the \spicedcontrolmore (median: 0.04). 
Similar findings are observed for the ordinal model (\Cref{fig:latex-figures-spiced-control--LessSciFi-MoreSciFi--wurl-et-al-2024-ordinal-aggr-score}).
%

\begin{figure}[tb]
    \centering
    \includegraphics[width=\linewidth]{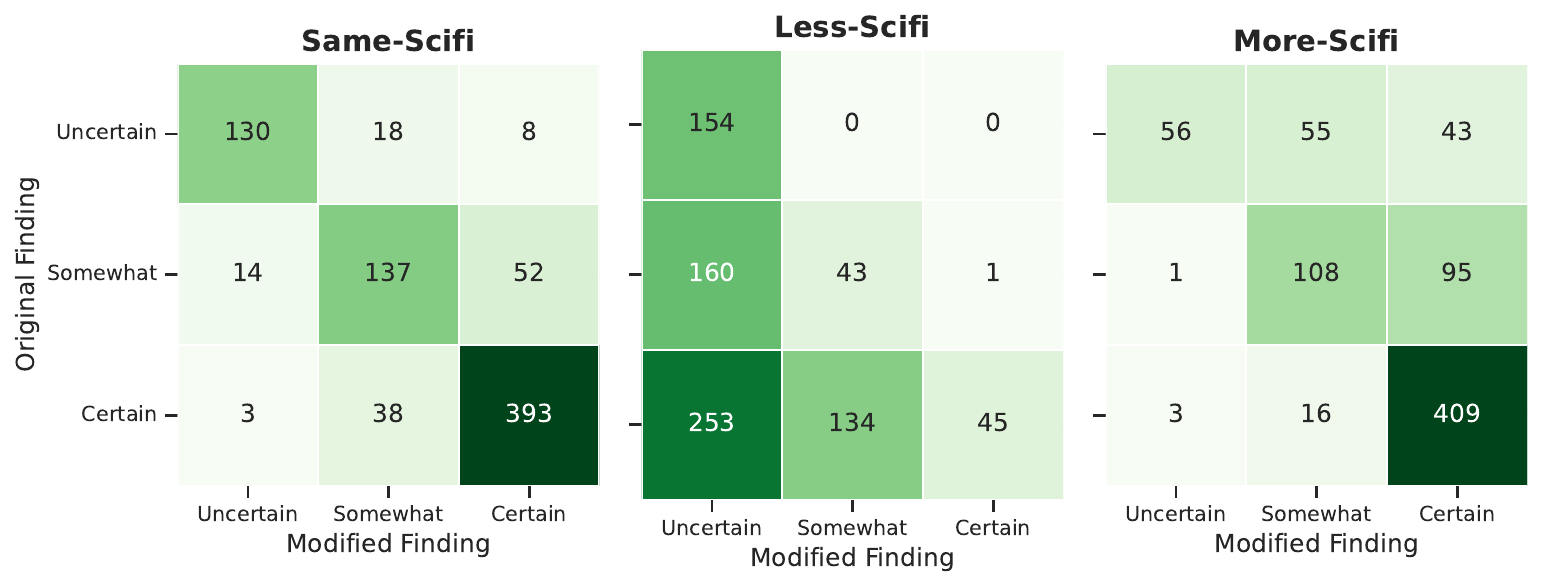} 
    \caption{Class predictions using the 3-class classifier proposed in~\citet{wuehrl-etal-2024-understanding}. Labels for original and modified finding are determined based on the \textit{maximum a posteriori} predictions of the models. 
    }
    \label{fig:latex-figures-spiced-control--wurl-et-al-2024-classif-conf-matrices}
\end{figure}

\paragraph{A.2.3.3. LM-as-a-Judge: Absolute Scoring.} 

Having considered multiple fine-tuned variants of BERT-based models, we now turn to LM-as-a-judge approaches~\citep{liu-etal-2023-g,gu2025surveyllmasajudge}. While prior work suggests the superiority of fine-tuned models~\citep{wuehrl-etal-2024-understanding}, we do not observe the same trend in our controlled datasets.

These differences can be attributed to both model capabilities and prompting strategies: prior work evaluates older-generation chat models, such as \texttt{Llama 2 (Chat-13B)} and \texttt{MiXtral (8x7B)}, and focuses on absolute scoring, whereas we use \texttt{gpt-5.4-mini} and instead emphasize pairwise comparisons. The following sections provide additional details on this approach.

\paragraph{Methodology.}
Motivated by the probabilistic perspective on certainty scoring, we first conduct a small-scale experiment to elicit absolute certainty scores in the range $[0, 1]$ using \texttt{gpt-5.4-nano-2026-03-17}(\texttt{temperature=0}; \texttt{max\_tokens=20}). This experiment is performed on 100 randomly sampled examples from each control dataset.

\paragraph{Results.} 
The results fall short of our expectations, as the differences in certainty scores are not significant, yielding a median $\Delta C \approx 0$. Upon manual inspection, we observe limitations similar to those of BERT-based models: \texttt{gpt-5.4-nano-2026-03-17} often assigns identical scores to sentences with subtle modifications, such as adding adverbs or adjectives that cause slight modifications in emphasis (\eg strongly, potentially, appear).

\paragraph{Analysis.} 
We attribute this behavior to the lack of anchoring in such approaches. Without an explicit reference point indicating what constitutes a higher level of certainty, the model tends to assign similar scores even when subtle differences exist. As a result, this \textit{reference-less} setting makes it difficult to capture fine-grained variations in certainty. In contrast, providing both sentences jointly may allow the model to better distinguish relative differences in certainty.

\paragraph{A.2.3.4. LM-as-a-Judge: Pairwise Comparison.}
Unlike the previous LM-as-a-judge approach, which assigns each text an absolute score and \textit{then} compares scores across paired findings, we now explore how directly comparing the certainty of two texts captures certainty shifts in the synthetic datasets.

\paragraph{Methodology.} 
We prompt \texttt{gpt-5.4-nano} and \texttt{gpt-5.4-mini} to compare the original finding with its modified version and identify the most certain sounding one.\footnote{Experiments are conducted throughout March 2026 using \texttt{gpt-5.4-nano-2026-03-17} and \texttt{gpt-5.4-mini-2026-03-17}.}
To mitigate potential positional biases, we prompt the LM twice, reversing the input order. Intuitively, a well-calibrated LM-as-a-judge should correctly discriminate between more certain and less certain text. For example, in \spicedcontrolmore, we expect the model to select the modified variant regardless of its position in the input. At the same time, the model should recognize when both texts express similar levels of certainty. In other words, for \spicedcontrolsame, we expect the model to be able to detect \textit{Ties}. 

Because we obtain two judge predictions for each pair, we combine them as follows.
Predictions that correctly reflect the input ordering (\eg predicting \textit{Text A} in the original order and \textit{Text B} when the order is reversed) or that are both ties (\eg \textit{Tie, Tie}) are considered consistent and used as the final predictions.
In contrast, if only one prediction is a \textit{Tie} (\eg \textit{*, Tie} or \textit{Tie, *}), or if the predictions remain the same regardless of input order (\eg \textit{Text A, Text A}), we classify them as \textit{Inconsistent}.%
\begin{figure}[bt]
\centering
\begin{tcolorbox}[fonttitle=\fontfamily{pbk}\selectfont\bfseries,fontupper=\fontfamily{ppl}\selectfont\itshape,fontlower=\fontfamily{put}\selectfont\scshape,
                  width=\linewidth,
                  arc=1mm, auto outer arc]
\begin{Verbatim}[fontsize=\footnotesize,breaklines=true, breaksymbol={}]
You have expertise in cognitive science and linguistics and specialize in detecting epistemic language. Epistemic language refers to expressions that convey uncertainty, belief, or degree of certainty (\eg "may", "might", "likely", "suggests", "appears").

Given a scientific finding, your task is to determine which of the two texts (if any) is communicated with more certainty. If both texts convey about the same certainty, consider it a tie. Certainty means having complete confidence in something without any doubts. Uncertainty is the opposite, referring to a state of doubt or lack of confidence about the proposition.
Before answering, think step by step about the certainty sources in each text and then provide your final decision.

[Requirements]
- Share your reasoning before the final answer.
- Enclose your final decision in <final_answer></final_answer> tags.
- The final decision should be 'Text A' if the certainty communicated in Text A is greater than that communicated in Text B. 
- The final decision should be 'Text B' if the certainty communicated in Text B is greater than that communicated in Text A. 
- The final decision should be 'Tie' if the certainty communicated in Text A and Text B is roughly the same.
\end{Verbatim}
\end{tcolorbox}
\caption{System prompt for determining pairwise certainty differences in two sentences. 
}
\label{fig:system-prompt:spiced-control-relative-ranking}
\end{figure}

\paragraph{Results.} 
Preliminary experiments comparing \texttt{gpt-5.4-nano} and \texttt{gpt-5.4-mini} on 100 randomly sampled examples from each control dataset show that \texttt{gpt-5.4-mini} more accurately captures directional certainty shifts (see \Cref{tab:llm-judge-pairwise-comparison}). In particular, \texttt{gpt-5.4-mini} is better able to distinguish the intended changes in \spicedcontrolless and \spicedcontrolmore, whereas \texttt{gpt-5.4-nano} exhibits a higher tendency to predict ties (+34\% in \spicedcontrolless and +10\% in \spicedcontrolmore), while maintaining a comparable tie rate in \spicedcontrolsame.
Furthermore, applying pairwise comparison with chain-of-thought reasoning (CoT) improves the performance of \texttt{gpt-5.4-mini} on \spicedcontrolmore and \spicedcontrolless, increasing accuracy by 11\% in both datasets. 
However, this is not the case for \spicedcontrolsame, where the number of pairs denoted \textit{Tie} decreases by 11 examples compared to no CoT.
\begin{table}[tb]
\small
\centering
\setlength{\tabcolsep}{6pt}
\renewcommand{\arraystretch}{1.2}
\begin{tabular}{l cccc}
\toprule
\textbf{Model} 
  & \textbf{Cert-$\uparrow$} 
  & \textbf{Cert-$\downarrow$} 
  & \textbf{Ties} 
  & \textbf{Incons} \\
\midrule
\rowcolor{sectionbg}
\multicolumn{5}{l}{\textit{Same Certainty}} \\ %
  gpt-5.4-nano         &  3 & 0 & 76 & 21 \\
  gpt-5.4-mini         &  4 & 0 & 73 & 23 \\
  gpt-5.4-mini+CoT           &  9 & 4 & 62 & 25 \\
\cmidrule(lr){1-5}
\rowcolor{sectionbg}
\multicolumn{5}{l}{\textit{Less Certainty}} \\ %
  gpt-5.4-nano      & 2 & 52 &  3 & 43 \\
  gpt-5.4-mini      & 0 & 88 &  1 & 11 \\
  gpt-5.4-mini+CoT        & 0 & 99 &  0 &  1 \\
\cmidrule(lr){1-5}
\rowcolor{sectionbg}
\multicolumn{5}{l}{\textit{More Certainty}} \\ %
  gpt-5.4-nano      & 69 &  1 &  5 & 25 \\
  gpt-5.4-mini      & 80 &  0 &  6 & 14 \\ 
  gpt-5.4-mini+CoT  & 91 &  0 &  3 &  6 \\
\bottomrule
\end{tabular}
\caption{Relative certainty differences under different pairwise comparison LLM-as-a-judge models. The results are reported over 100 random examples from 3 different sets with directional shifts. \textit{Cert-$\uparrow$}, \textit{Cert-$\downarrow$}, \textit{Ties} concern the number of pairs whose certainty increases, decreases, or remains the same, respectively. \textit{Incons} refers to the number of examples for which the LLM-as-a-judge produces inconsistent labels.}
\label{tab:llm-judge-pairwise-comparison}
\end{table}

\paragraph{Analysis.} 
Unlike \spicedcontrolmore and \spicedcontrolless, predictions for \spicedcontrolsame are more frequently labeled as \textit{Inconsistent}, a pattern that aligns with the higher annotator disagreement observed in \Cref{appdx:validation-uncertainty-estimator:dataset-validation}. Since these pairs are constructed to differ only marginally in communicated certainty, they are generally expected to be labeled as \textit{Tie}. However, the LLM judge often produces inconsistent predictions.
To better understand this behavior, we manually inspect more than 60 inconsistent examples and their reasoning traces. Although the LLM judge often acknowledges that the differences between texts are minimal, it tends to over-analyze these variations and force a decision in favor of one text. For instance, when the judge outputs order-invariant labels such as (\textit{Text A}, \textit{Text A}) or (\textit{Text B}, \textit{Text B}), the accompanying explanations frequently contain contradictory arguments, with similar reasoning used to justify opposing conclusions. In other cases, the judge correctly identifies a tie in one direction but fails to do so when the order is reversed, even when differences are minimal (\eg ``may serve'' vs. ``could function as,'' or active vs. passive voice).
We hypothesize that, in the presence of such subtle differences, the LLM judge becomes more susceptible to biases such as position bias and confirmation bias~\citep{zhao2021calibrate,gudibande2024the}, leading it to over-reason about negligible variations and produce inconsistent outcomes. Additionally, task ambiguity may further contribute to this behavior, as we do not explicitly define a threshold for what constitutes a meaningful difference in relative certainty. In the absence of such guidance, the model may apply inconsistent internal criteria when deciding between \textit{Tie} and directional labels.

\section{Human Experiment Details}
\label{appdx:sec:details-user-studies}

\begin{figure*}[tb]
    \centering
    \includegraphics[width=1\linewidth]{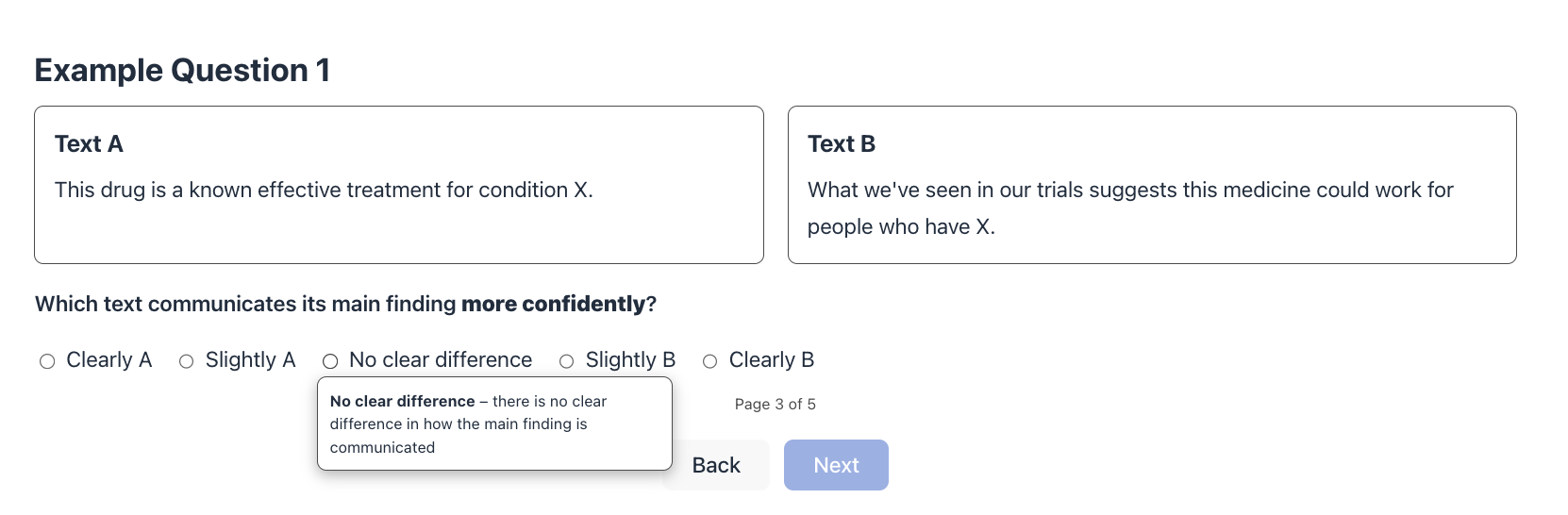}
    \caption{Participants' view of the annotation Interface. Given a pair of scientific findings discussing the same main claim, the participants must select which of the two (Text A or Text B) communicates the main claim more confidently. The selection is made using a 5-point Likert Scale where the middle option indicates no significant difference.}
    \label{fig:participant-ui}
\end{figure*}

\paragraph{Task description.} 
We designed an annotation task in which participants served as human judges and compared two texts, \textit{Text A} and \textit{Text B}, that expressed the same underlying claim, see \Cref{fig:participant-ui}. The two texts differed only in style: one was a human-written scientific finding and the other was LLM-generated (instructed to output either a paraphrase or a version suitable for a journal). 
Participants were not informed of this distinction. For each item, participants indicated which text communicated the claim more confidently using a 5-point Likert scale. 
Each participant annotated 20 items. 
The order of items and the left-right placement of \textit{Text A} and \textit{Text B} were randomized to mitigate order effects.

\paragraph{Procedures.} The study was conducted online via Prolific. After providing consent, participants reviewed the study instructions and completed five screening checks. To proceed to the main study, participants had to select the majority-preferred choice in at least 4 of the 5 screening examples. Those who passed the screening checks then completed 20 annotation questions using the interface shown in Figure~\ref{fig:participant-ui}. An additional attention check was embedded among the annotation questions to ensure data quality. After completing the main task, participants filled out a short post-study survey. The survey responses are not used in the analysis; the survey was originally included during piloting to refine the interface and study design and was retained in the main study for completeness.

\paragraph{Participants.} We recruited 138 U.S.-based, English-speaking adults aged 18 or older, each with at least an undergraduate degree and a Prolific approval rating of at least 97\%. We excluded 9 annotators based on pre-specified quality checks: 3 for excessive off-screen time, 1 for fast straight-lining responses, and 5 for failing the attention check. The final sample comprised 129 participants.

Participants had a mean age of 39.6 years (SD = 12.0); 57 identified as male, 71 as female, and 1 preferred not to disclose. In terms of education, 92 had an undergraduate degree, and 37 had a master’s degree or higher. Participants received \$4 for participation (median duration: 15 minutes). The experiment received Institutional Review Board (IRB) approval.\footnote{
The study was approved by the Institutional Review Board (IRB) at the University of California, Irvine
(protocol \#1493).
}

\subsection{Additional Results}
\label{apdx:additional-human-experiment-results}

\begin{figure}[tb]
    \centering
    \includegraphics[width=\linewidth]{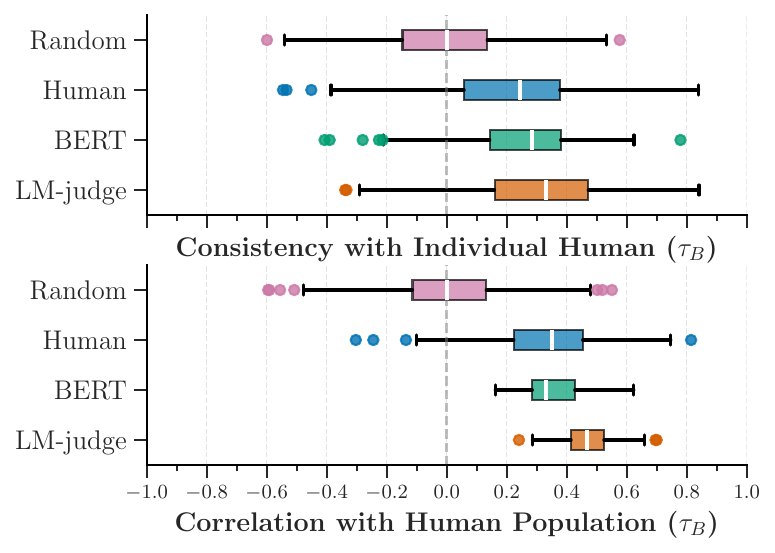}
    \caption{
    Correlation of different \certaintydistortion estimators with the mean prediction of human-population, measured using Kendall $\tau_B$ correlations. 
    BERT refers to the 3-way classification BERT-based estimator proposed in~\citet{wuehrl-etal-2024-understanding}. 
    We find that LM-as-a-judge approach (parameterized by \evalgpt) is as consistent with humans as randomly selected Human-Human pair (top), and achieves better average correlation with the human population-level consensus (bottom).
    }
    \label{fig:human-validation:spiced:paraphrase-both-metrics}
\end{figure}

To complement the analysis in the main paper, we also compare each estimator with the average consistency with individuals in the human population. 
These results are presented in \Cref{fig:human-validation:spiced:paraphrase-both-metrics}.  We find that LM-as-a-judge approach (parameterized by \evalgpt) is as consistent with humans as a randomly selected Human-Human pair (top), and achieves better average correlation with the human population-level consensus (bottom). 

\begin{figure}[tb]
    \centering
    \includegraphics[width=\linewidth]{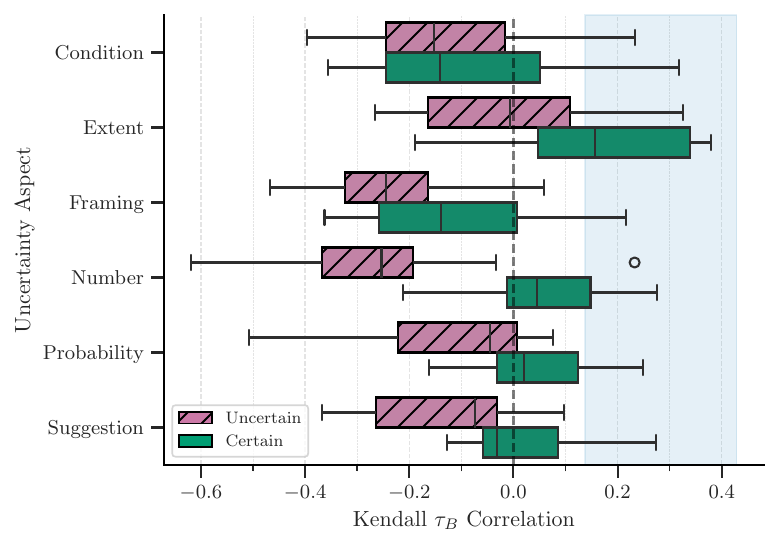} 
    \caption{
    Consistency of the aspect-level classifiers proposed by \citet{pei-jurgens-2021-measuring} with individual human annotators. Consistency is measured using Kendall’s $\tau_B$ correlation coefficient. \textit{Certain} and \textit{Uncertain} correspond to computing certainty distortion using the difference in the predicted probabilities of the Certain and Uncertain classes, respectively. To facilitate comparison with average human-to-human correlations, the 25\%–75\% range of human correlation coefficients is shown as a shaded region.
    }
    \label{fig:human-study:consistency-humans:all-bert-based-pei-jurgens}
\end{figure}

In the main paper, we only report the correlation values for the BERT-based metric proposed by~\citet{wuehrl-etal-2024-understanding}. This section summarizes the correlations obtained for an extended list of classifiers. 
\Cref{fig:human-study:consistency-humans:all-bert-based-pei-jurgens} shows the correlation between the models proposed by \citet{pei-jurgens-2021-measuring} and human consensus. The models were fine-tuned to predict a certainty score for each sentence. Specifically, we test the 3-class classifiers that model six uncertainty aspects: Condition, Extent, Framing, Numerical, Probability, and Suggestion. 
Because these models have not been fine-tuned for pairwise comparisons, we induce a relative ordering by comparing their class scores. Overall, most classifiers exhibit negative or near-zero correlation with human annotators, suggesting behavior that is less consistent than that of a randomly selected individual. Although the Extent (Certain) classifier overlaps with the annotator-to-annotator distribution (blue shaded region), the variance in its behavior is too large for it to serve as a reliable proxy for human annotation in a pairwise setting.
For more details on the differences between these models, we refer you to Appendix \ref{apdx:controlled-eval}. 

\begin{figure}[tb]
    \centering
    \includegraphics[width=\linewidth]{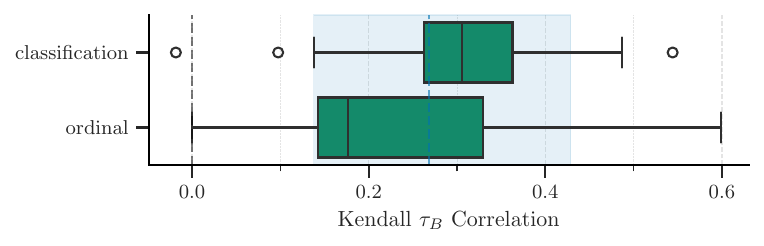} 
    \caption{
    Consistency of the aspect-level classifiers proposed by \citet{wuehrl-etal-2024-understanding} with individual human annotators. Consistency is measured using Kendall’s $\tau_B$ correlation coefficient. To facilitate comparison with average human-to-human correlations, the 25\%–75\% range of human correlation coefficients is shown as a shaded region. The median of the human behavior is also included  (blue dashed line) for comparison.
    }
    \label{fig:human-study:consistency-humans:all-bert-based-wuerl-et-al-2024}
\end{figure}

\Cref{fig:human-study:consistency-humans:all-bert-based-wuerl-et-al-2024} illustrates the results for the classifiers trained using the data collected in \citet{wuehrl-etal-2024-understanding}. 
They propose to train an LM using an ordinal 4-point Likert scale (1) Uncertain, (2) Somewhat Uncertain, (3) Somewhat Certain, and (4) Certain. The data is collected by asking participants to determine whether a sentence expresses uncertainty---irrespective of whether it arises from framing, numerical, extent, or other types of uncertainty. 
The final dataset exhibits severe class imbalance which the authors address by merging the two uncertain classes (labels 1-2), resulting in 20\% uncertain labels. 
Because the original models were not released and the details of the loss function are unclear, we fine-tune two variants ourselves: a 3-class classifier and an ordinal regression model.
The ordinal regression formulation behaves similarly to a 3-way classifier but imposes a monotonic ordering over the classes, which we expect to encourage faster learning and produce more calibrated certainty signals.
Results show that both variants overlap with the human distribution, with the classifier model (\textit{classifier}) achieving a median Kendall $\tau_B$ correlation higher than that of a randomly selected human participant. Still, these classifiers continue to underperform relative to the LLM-as-a-judge approach discussed in the main paper.
Overall, these results provide additional justification for adopting an LLM-as-a-judge framework, given its stronger alignment with human judgments and improved consistency.

\section{Prompts}
\label{apdx:prompts}

This section lists all prompts used in the paper.
Appendix~\ref{apdx:ssec:task-prompts} lists the task-specific prompts used for  generation.

\subsection{Task Prompts}
\label{apdx:ssec:task-prompts}

\begin{itemize}
    \item  Single-sentence level prompts: \begin{itemize}
            \item Paraphrasing - SPICED: \Cref{fig:paraphrasing-prompt-spiced}
            \item News Rewriting - SPICED: \Cref{fig:rewrite-prompt-news}
            \item Paraphrasing - MIMIC: \Cref{fig:paraphrasing-prompt-medical-sent-level}
            \item Simplification - MIMIC: \Cref{fig:simpl-prompt-medical-sent-level}
        \end{itemize}

    \item Multi-sentence level prompts: \begin{itemize}
        \item Paraphrasing - Academic Papers: \Cref{fig:paraphrase-prompt-academic-doc-level}
        \item Summarization - Academic Papers:  \Cref{fig:summarize-prompt-academic-doc-level}
        \item Paraphrasing - MIMIC: \Cref{fig:paraphrasing-prompt-medical-doc-level}
        \item Simplification - MIMIC: \Cref{fig:simpli-prompt-medical-doc-level}
    \end{itemize}
\end{itemize}

%
%
\begin{figure}[tbhp]
\centering
\begin{tcolorbox}[fonttitle=\fontfamily{pbk},fontupper=\fontfamily{ppl}\selectfont\itshape,fontlower=\fontfamily{put}\selectfont\scshape,
                  width=\linewidth,
                  arc=1mm, auto outer arc]
\begin{Verbatim}[fontsize=\footnotesize,breaklines=true, breaksymbol={}]
Your task is to paraphrase an English text. When paraphrasing, avoid copying phrases from the provided text. Your paraphrase must be 1 sentence long. Return **only** the paraphrased text.
\end{Verbatim}
\end{tcolorbox}
\caption{System prompt used in the sentence-level paraphrasing task (SPICED dataset).}
\label{fig:paraphrasing-prompt-spiced}
\end{figure}
\begin{figure}[tbhp]
\centering
\begin{tcolorbox}[fonttitle=\fontfamily{pbk},fontupper=\fontfamily{ppl}\selectfont\itshape,fontlower=\fontfamily{put}\selectfont\scshape,
                  width=\linewidth,
                  arc=1mm, auto outer arc]
\begin{Verbatim}[fontsize=\footnotesize,breaklines=true, breaksymbol={}]
Imagine you are a New York Times journalist. Rewrite the following scientific finding as **one sentence** suitable for publication. Return **only** the sentence.
\end{Verbatim}
\end{tcolorbox}
\caption{System prompt used in the sentence-level News Rewriting task (SPICED dataset).}
\label{fig:rewrite-prompt-news}
\end{figure}
\begin{figure}[tbhp]
\centering
\begin{tcolorbox}[fonttitle=\fontfamily{pbk},fontupper=\fontfamily{ppl}\selectfont\itshape,fontlower=\fontfamily{put}\selectfont\scshape,
                  width=\linewidth,
                  arc=1mm, auto outer arc]
\begin{Verbatim}[fontsize=\footnotesize,breaklines=true, breaksymbol={}]
Paraphrase this radiology findings sentence. Your paraphrase must be 1 sentence long. Return **only** the paraphrased text.
\end{Verbatim}
\end{tcolorbox}
\caption{System prompt used in the sentence-level Paraphrasing task (MIMIC-CXR dataset).}
\label{fig:paraphrasing-prompt-medical-sent-level}
\end{figure}
\begin{figure}[tbhp]
\centering
\begin{tcolorbox}[fonttitle=\fontfamily{pbk},fontupper=\fontfamily{ppl}\selectfont\itshape,fontlower=\fontfamily{put}\selectfont\scshape,
                  width=\linewidth,
                  arc=1mm, auto outer arc]
\begin{Verbatim}[fontsize=\footnotesize,breaklines=true, breaksymbol={}]
You are simplifying a single sentence from the Findings section of a radiology report.


Instructions:
- Rewrite the sentence in simpler and clearer language.
- Preserve all clinical information.
- Do not add new findings or interpretations.
- Do not omit medically important details.
- Keep the output as a single sentence.
- Use concise and natural wording.
- Avoid unnecessary medical jargon when possible.
\end{Verbatim}
\end{tcolorbox}
\caption{System prompt used in the sentence-level simplification task (MIMIC-CXR dataset).}
\label{fig:simpl-prompt-medical-sent-level}
\end{figure}
\begin{figure}[tbhp]
\centering
\begin{tcolorbox}[fonttitle=\fontfamily{pbk},fontupper=\fontfamily{ppl}\selectfont\itshape,fontlower=\fontfamily{put}\selectfont\scshape,
                  width=\linewidth,
                  arc=1mm, auto outer arc]
\begin{Verbatim}[fontsize=\footnotesize,breaklines=true, breaksymbol={}]
You have experience in academic writing.

Your task is to rephrase the following abstract in English.

Input text to process:
"""{text}"""

Requirements:
 - Preserve the academic tone.
 - The rewritten version should be roughly comparable in length to the original, around {target_len} words.
 - Do not include any introductory sentence such as "Here is a rephrased version..."
 - Do not include explanations, notes, labels, bullet points, or comments.
 - Write **only** the rewritten abstract text.
\end{Verbatim}
\end{tcolorbox}
\caption{System prompt used in the document-level paraphrasing task (Academic Papers dataset).}
\label{fig:paraphrase-prompt-academic-doc-level}
\end{figure}
\begin{figure}[tbhp]
\centering
\begin{tcolorbox}[fonttitle=\fontfamily{pbk},fontupper=\fontfamily{ppl}\selectfont\itshape,fontlower=\fontfamily{put}\selectfont\scshape,
                  width=\linewidth,
                  arc=1mm, auto outer arc]
\begin{Verbatim}[fontsize=\footnotesize,breaklines=true, breaksymbol={}]
You have experience in academic writing. You are doing a literature review and want to summarize the main findings of academic abstracts.

Your task is to summarize the following academic abstract in English in a single sentence.

Input text to process:
"""{text}"""

Requirements:
- Focus on the main finding or contribution of the abstract.
- Include the research question, data/method, or setting only if needed to understand the finding.
- Preserve the academic tone.
- Write exactly 1 sentence.
- Do not include any introductory sentence such as "Here is a summary..."
- Do not include explanations, notes, labels, bullet points, or comments.
- Write **only** the summary sentence.
\end{Verbatim}
\end{tcolorbox}
\caption{User prompt used in the document-level summarization task (Academic Papers dataset).}
\label{fig:summarize-prompt-academic-doc-level}
\end{figure}
\begin{figure}[tbhp]
\centering
\begin{tcolorbox}[fonttitle=\fontfamily{pbk},fontupper=\fontfamily{ppl}\selectfont\itshape,fontlower=\fontfamily{put}\selectfont\scshape,
                  width=\linewidth,
                  arc=1mm, auto outer arc]
\begin{Verbatim}[fontsize=\footnotesize,breaklines=true, breaksymbol={}]
Your task is to paraphrase an English text. When paraphrasing, avoid copying phrases from the provided text. The paraphrase should be no longer than {n_sentences} sentences. Return **only** the paraphrased text. Do not include a header, preamble, or any text outside the paraphrased text.
\end{Verbatim}
\end{tcolorbox}
\caption{System prompt used in the document-level paraphrasing task (MIMIC-CXR dataset). \texttt{n\_sentences} is a placeholder value that is replaced during generation by the number of sentences of each Findings section. This parameter is a way of controlling for length.}
\label{fig:paraphrasing-prompt-medical-doc-level}
\end{figure}
\begin{figure}[tbhp]
\centering
\begin{tcolorbox}[fonttitle=\fontfamily{pbk},fontupper=\fontfamily{ppl}\selectfont\itshape,fontlower=\fontfamily{put}\selectfont\scshape,
                  width=\linewidth,
                  arc=1mm, auto outer arc]
\begin{Verbatim}[fontsize=\footnotesize,breaklines=true, breaksymbol={}]
You are explaining part of a chest X-ray report to a patient who has no medical background.

Rewrite the Findings below in simple, plain language a patient can understand.

Format requirements:
- Return a numbered list (1., 2., 3., ...).
- Each item must be a single sentence that explains one radiological observation from the Findings.
- Include between 1 and 8 items, ordered as they appear in the Findings.
- Use everyday language; avoid medical jargon. If a medical term is unavoidable, briefly explain it.
- Do not add reassurance, recommendations, or any information not present in the Findings.
- Do not include a header, preamble, or any text outside the list.
\end{Verbatim}
\end{tcolorbox}
\caption{System prompt used in the document-level simplification task (MIMIC-CXR dataset).}
\label{fig:simpli-prompt-medical-doc-level}
\end{figure}

\subsection{Mitigation Instructions}
\label{apdx:ssec:mitigation-prompts}

To each of the task prompts above, we append the following: 
\begin{itemize}
    \item \textit{Preserve semantics}: ``Preserve semantics of the original text.''
    \item \textit{Do not hallucinate}: ``Preserve semantics of the original text. Do not hallucinate.''
    \item \textit{Preserve Certainty}: ``Preserve semantics of the original text. Do not hallucinate. Preserve the degree of certainty in the original text.''
\end{itemize}

\subsection{Evaluation Prompts}
\label{apdx:ssec:eval-prompts}

The final single-sentence evaluation prompt is listed in \Cref{fig:prompt:llm-judge-5-options-structured}. It resembles the prompt provided to the humans, consisting of two demonstrations with clear explanations.
\begin{figure*}[tb]
\centering
\begin{tcolorbox}[
fonttitle=\fontfamily{pbk}\selectfont\bfseries,
    fontupper=\fontfamily{ppl}\selectfont\itshape,
    fontlower=\fontfamily{put}\selectfont\scshape,
    width=\linewidth,
    arc=1mm, auto outer arc
]
\begin{Verbatim}[fontsize=\footnotesize,breaklines=true, breaksymbol={}]
You have expertise in cognitive science and linguistics and specialize in detecting epistemic language. Epistemic language refers to expressions that convey uncertainty, belief, or degree of certainty (\eg "may", "might", "likely", "suggests", "appears"). In this study, you will compare two versions of the same text and judge which presents its main finding more confidently. 

Here are two representative examples to help you familiarize with the task.

## Example 1.
Example input 1:
Text A: This drug is a known effective treatment for condition X.
Text B: What we've seen in our trials suggests this medicine could work for people who have X.
Question: Which text communicates its main finding **more confidently**?
Options: Clearly A, Slightly A, No Clear Difference, Slightly B, Clearly B

Example output 1:
<explanation>Text A uses assertive, declarative language (“is,” “known”), presenting the claim as established fact with no uncertainty. Text B uses hedging terms like “suggests” and “could,” with some users also mentioning that the claim is presented as a possibility based on limited trial evidence. Overall, Text A expresses higher confidence than Text B.</explanation>
<final_answer>Clearly A</final_answer>

## Example 2.
Example Input 2:
Text A: This drug is a known effective treatment for condition X.
Text B: Evidence demonstrates the efficacy of this therapeutic agent in treating condition X.
Question: Which text communicates its main finding **more confidently**?
Options: Clearly A, Slightly A, No Clear Difference, Slightly B, Clearly B

Example output 2:
<explanation>Text A uses assertive, declarative language (“is,” “known”), presenting the claim as established fact with no uncertainty. Text B uses high-certainty phrasing (“demonstrates efficacy”), which similarly conveys the claim as proven. Overall, both texts express the relationship as a confirmed reality, with no meaningful difference in certainty.</explanation>
<final_answer>No Clear Difference</final_answer>

## Requirements
- When the user provides two texts discussing the same main finding, determine which text communicates the main finding more confidently. 
- Think step by step about which of the two texts is more certain, then conclude by selecting one of the provided options as the final answer.
\end{Verbatim}
\end{tcolorbox}
\caption{System prompt for determining pairwise certainty differences in two texts discussing the same main finding. Using \evalgpt (temperature=0), we find this prompt to exhibit good alignment with the directional shifts in the controlled datasets (validated in Appendix \ref{appdx:validation-uncertainty-estimator:dataset-validation}) as well as with overall statistics of humans' perception of certainty (validated in \Cref{ssec:human-validation}).}
\label{fig:prompt:llm-judge-5-options-structured}
\end{figure*}

\section{Implementation Details}
\label{appdx:additional-method-details}

All experiments were run using inference APIs throughout March-May 2026. 
Generator models (\gengpt, \gengemini, \genllama, \genqwen, \genclaude) use Chat Completions API provided through the OpenAI SDK. 
The evaluator model (\evalgpt) leverages the Responses API. 
Specifically, we use the structured outputs feature to ensure the LM always generates an explanation and the final answer, which is one of the answer choices ``Clearly A'', ``Slightly A'', ``No Clear Difference'', ``Slightly B'', and ``Clearly B''.

\subsection{LM-as-a-judge: Response Aggregation}
\label{apdx:additional-method-details-llm-judge}

The system prompt used for the LLM-as-a-judge evaluation is provided in \Cref{fig:prompt:llm-judge-5-options-structured}. It mirrors the instructions provided to human annotators, including two demonstrations and a 5-point Likert scale. To ensure reproducibility, we decode the judge's responses greedily (\texttt{T=0}), eliminating stochasticity in its outputs.

When querying \evalgpt, we collect two responses per example (one for each item ordering) and compare them to assess consistency. We define a response pair as \textit{consistent} if both orderings yield the same label, and \textit{inconsistent} otherwise. For near-consistent cases where both responses rank the same text as more certain but at differing magnitudes (\eg ``Slightly Original'' vs. ``Clearly Original''), we apply a conservative resolution rule, taking the weaker of the two labels as the final annotation.
We find that between 7.3\% (\gengpt) and 12.6\% (\gengemini) of examples yield conflicting labels across the two orderings in paraphrase (SPICED) task, and between 10.1\% (\gengpt) and 18.1\% (\gengemini, \genllama) in news (SPICED) task. 
Manual inspection of inconsistent reasoning chains reveals two recurring patterns. First, when the certainty difference between texts is subtle, the model produces similar reasoning under both orderings but favors the opposite item, suggesting sensitivity to surface-level text properties rather than fundamental disagreement. 
Second, we observe a positional asymmetry: the judge tends to reason more thoroughly about the second-listed item (Text B), occasionally surfacing epistemic cues not noted for Text A (\eg contingent claims, speaker commitment).
To address these inconsistencies, we introduce a moderator model---a stronger LM-judge that reviews both reasoning chains and produces its own reasoning and final label. To guard against positional bias in the moderator's responses, we also query it under both orderings and apply the same aggregation procedure described before.

Using \gptbig as a moderator reduces the inconsistency rate from 10\% to 6.6\% on paraphrase (SPICED) and from 15\% to 9.3\% on news (SPICED). Among resolved inconsistencies, 50\% are arbitrated as Ties, 11.3\% as a decrease in certainty, and 4.2\% as an increase in certainty. 
The moderator fails to resolve the remaining 33.9\% of inconsistent examples. 
Given that even a stronger model cannot adjudicate these examples, \textbf{\textit{we treat them as genuinely ambiguous comparisons}}.
Inspection of the moderator's reasoning chains on these residual cases reveals that the moderator exhibits the same failure mode as the base judge: it produces detailed analyses of verbal framing and epistemic markers, but arrives at opposite conclusions depending on item ordering, suggesting the \textbf{\textit{ambiguity is a property of the examples themselves rather than a limitation of the evaluation model}}. 

\subsection{LM-as-a-judge: Certainty of Source Text}
\label{apdx:additional-method-details-llm-judge:bradley-terry}

In \Cref{ssec:sentence-level-distortion-rates}, we analyze how certainty 
distortion varies across different levels of source-text certainty. To assign 
certainty labels, we use a probabilistic framework to infer a global ranking 
from pairwise comparisons between source texts, then divide the ranked examples 
into \textit{low}, \textit{medium}, and \textit{high} certainty groups using a 
25/50/25 split. For SPICED ($N=397$), this yields 100 \textit{low}, 197 
\textit{medium}, and 100 \textit{high} examples; an analogous procedure is 
applied to the MIMIC-CXR sentence-level dataset.

\paragraph{Bradley--Terry Model with Ties.}
To infer a latent certainty score for each source text, we fit a 
Bradley--Terry model with ties \citep{rao1967ties} to a dataset of pairwise 
comparisons. The model assigns each source text $x_i$ a latent strength 
parameter $\lambda_i > 0$ and introduces a tie-threshold parameter $\theta 
\geq 1$. Given a pair $(x_i, x_j)$, the probability that $x_i$ is judged 
more certain is
\begin{equation}
    P(x_i \succ x_j) = \frac{\lambda_i}{\lambda_i + \theta\sqrt{\lambda_i \lambda_j}},
\end{equation}
which increases with $\lambda_i$ and decreases with $\lambda_j$, as expected. 
The probability of a tie is
\begin{equation}
    P(x_i \sim x_j) = \frac{(\theta^2 - 1)\sqrt{\lambda_i \lambda_j}}
                            {(\lambda_i + \theta\sqrt{\lambda_i \lambda_j})
                             (\theta\sqrt{\lambda_i \lambda_j} + \lambda_j)},
\end{equation}
which vanishes as $\theta \to 1$ (recovering standard Bradley--Terry~\citep{BradleyTerry1952}) and grows as $\theta$ increases, reflecting a greater propensity for tied judgments. We estimate $\{\lambda_i\}$ and $\theta$ jointly by maximising 
the log-likelihood of the observed pairwise judgments via the iterative MLE 
procedure of \citet{rao1967ties}, and use the resulting $\{\lambda_i\}$ to rank all source texts.

\paragraph{Collecting Pairwise Comparisons.}
We collect $\mathcal{D}_{P}$ in two passes using \evalgpt and the prompt in 
\Cref{fig:prompt:llm-judge-5-options-structured}. 
To reduce positional bias, every pair is evaluated in both presentation orders. We use a temperature of \texttt{T=0.7} throughout, so that the aggregated counts $(w_i, w_j, t)$ reflect both the direction and the consistency of the model's preferences 
across repeated judgments.

\textit{Pass 1: Global ranking.} Each source text $x_i$ is randomly paired with $k$ other examples and each pair receives two judgments (one corresponding to each order). 
To limit costs, we use $k=25$ for SPICED and $k=10$ for MIMIC-CXR, since the latter is approximately twice as large; in both cases $k \geq \log N$, ensuring the pairwise comparison graph 
is connected with high probability. 
Although larger $k$ would improve ranking quality, collecting $k \times N$ judgments is computationally costly, so we 
refine the initial ranking with a targeted second pass.

\textit{Pass 2: Fine-grained disambiguation.} 
After obtaining an initial ranking from Pass~1, we focus additional comparisons on examples that are close in rank, where pairwise differences are most ambiguous. When sampling 
comparison partners for $x_i$ (at rank $i$), we draw from a centered window $\{x_{i-w}, \ldots, x_{i-1}, x_{i+1}, \ldots, x_{i+w}\}$ rather than the full dataset. For each pair we now report 3 votes (totaling 6 judgments per pair). We use $k=5$ for SPICED and $k=8$ for MIMIC-CXR dataset. For each dataset, all comparison data from both passes are pooled into $\mathcal{D}_{P}$ before fitting the final model. 

\paragraph{Qualitative Examples.}

\Cref{tab:examples:btt-spiced,tab:examples:btt-mimic-cxr} provide qualitative examples for SPICED and MIMIC-CXR datasets, respectively. 
These are representative examples from the global certainty ranking, with the 5 top, 5 middle, and 5 bottom examples representing randomly selected high, medium, and low certainty examples.
\begin{table}[p]
\small
\centering
\resizebox{\linewidth}{!}{
\begin{tabular}{p{0.24\linewidth} p{0.73\linewidth}}
\toprule
\textbf{Strength ($\lambda$)} & \textbf{Source Text} \\
\midrule
\rowcolor{gray!10} 6.26 & RNA sequencing expression profiling revealed that expression of 456 identified genes was significantly changed in the tumor due to walnut consumption. \\
6.08 &	The biggest spike in Monterrey (7,027 tweets) occurred on August 25, 2011, the day of an attack at a casino where 53 people were killed. \\
\rowcolor{gray!10}4.71 & 	Buffering against acidification or inhibiting lactic acid production fully rescues circadian oscillation. \\
4.32 & The current study is the first to demonstrate that HS students are using e-cigarettes to vaporize cannabis.\\
\rowcolor{gray!10}4.03 &	A woman who has large lips, suggesting a strong mating potential, with average length and width ratios will always be more attractive than a woman with narrow lips and average length and width ratios. \\
\addlinespace
0.6690 & This highlights a consistent neural substrate for each component of empathy. \\
\rowcolor{gray!10}0.2742 & Non-linear associations with diabetes were observed for age at menarche, parity and age at first birth.\\
0.1890 & This discovery supports colonial nesting behavior and potential nesting site fidelity in the Pterosauria.\\
\rowcolor{gray!10}0.1654 & Although absolute risks for neuropathy are low, CD is a potentially treatable condition with a young age of onset. \\
0.1505 & Surprisingly, in absence of any stimuli, some adherent cells closely resembling neuronal and glial morphology were also observed.\\
\addlinespace
\rowcolor{gray!10}0.0048 & Reduction in the incidence of colorectal cancer \textbf{may} influence the overall benefit-risk profile of aspirin.\\
0.0025 & It is \textbf{possible} that alcohol \textbf{may} only affect perception of attractiveness at a higher level of alcohol consumption, which was not observable with the naturalistic design used here.\\
\rowcolor{gray!10}0.0021 & Together, our results \textbf{suggest} that pro-resolution mechanisms may act in concert with increased cytotoxic immunity to reduce the growth of tumors in response to stretching.\\
0.0018 & These results \textbf{suggest} that processing juveniles in the adult system \textbf{may} not uniformly increase offending and \textbf{may} reduce offending in some circumstances.\\
\rowcolor{gray!10}0.0014 & These models \textbf{suggest} opossum expansion \textbf{may} be facilitated by agricultural land development and at the same time be limited by their inability to forage in snow.\\
\bottomrule
\end{tabular}}
\caption{Examples from the global certainty ranking for SPICED, estimated via the Bradley--Terry-with-ties model \citep{rao1967ties} with \evalgpt as the judge. Higher-ranked texts report direct observations or measurements, while lower-ranked texts draw interpretive conclusions hedged with markers such as \textit{may}, \textit{suggest}, and \textit{possible} (bold).}
\label{tab:examples:btt-spiced}
\end{table}
\begin{table}[tb]
\small
\centering
\resizebox{\linewidth}{!}{
\begin{tabular}{p{0.24\linewidth} p{0.73\linewidth}}
\toprule
\textbf{Strength ($\lambda$)} & \textbf{Source Text}  \\
\midrule 
\rowcolor{gray!10} 8.0	& Right infusion port catheter terminates in the low SVC. \\
7.6	& The lungs are well inflated and clear. \\
\rowcolor{gray!10}6.7 & PA and lateral views of the chest provided demonstrate clear well-expanded lungs without focal consolidation, effusion or pneumothorax. \\ 
5.7 & There is no pleural effusion, pneumothorax, or pulmonary edema. \\ 
\rowcolor{gray!10}5.1 & 	The right lung is clear. \\ 
\addlinespace
1.32 & Lung volumes are mildly decreased. \\
\rowcolor{gray!10}1.16 & A frontal semi-upright view of the chest was obtained portably. \\
1.07 &As compared to the previous radiograph, the pre-existing right upper and lateral peripheral consolidation has decreased in extent and severity. \\
\rowcolor{gray!10}0.32 & Lung apices are not well visualized.\\
0.12 & This \textbf{could} represent fluctuating atelectasis.\\
\addlinespace
\rowcolor{gray!10}0.00472 & \textbf{Some} focal opacity in the right upper zone laterally is noted -- \textbf{if} real, this is new, but it \textbf{could} represent artifact due to the presence of the subcutaneous emphysema. \\
0.00044 & This \textbf{could} \textbf{potentially} be projectional. \\
\rowcolor{gray!10}0.00038 & The small rounded indentation is seen on the gastric bubble, which \textbf{could} \textbf{possibly} represent a mass in the hernia. \\
0.00015 & There is obscuration of the left hemidiaphragm, which \textbf{could} be due to atelectasis, which \textbf{may} be due to atelectasis and low lung volumes. \\
\bottomrule
\end{tabular}}
\caption{Examples from the global certainty ranking for MIMIC-CXR, estimated via the Bradley--Terry-with-ties model \citep{rao1967ties} with \evalgpt as the judge. Higher-ranked texts express unambiguous clinical findings, while lower-ranked texts are hedged with epistemic markers such as \textit{could}, \textit{may}, and \textit{possibly} (bold).}
\label{tab:examples:btt-mimic-cxr}
\end{table}

\section{Datasets}
\label{apdx:sec:additional-dataset-details}

\textbf{SPICED} is a dataset about scientific communication~\citep{wright-etal-2022-SPICED}.
For each scientific finding, the dataset provides a news version and a twitter version, as well as a final information matching score that represents how likely these texts are to be discussing the same finding. These scores were obtained through human studies by collecting annotations for 5-point Likert scales, and then were averaged to create the final score. 
Although we do not use the paired finding, we restrict our analysis to the subset of single-sentence scientific findings with a paired news version whose score is equal or greater than 4.
This leads us to 397 total findings spanning four different categories, amongst which computer science, biology, psychology.

\textbf{Academic Abstracts} is a dataset that we extracted from journal publications in 2020 from Dimensions---a comprehensive research knowledge system developed by Digital Science.\footnote{\url{https://dimensions.ai.}} 
We focus on 2020 to ensure that the abstracts are, to the best of our knowledge, entirely human-written. We access the database using the licensed Dimensions API. 
We randomly sample 100 publications from five fields: Psychology (41k), Health Sciences (92k), Biological Sciences (160k), Biomedical and Clinical Sciences (447k), and Earth \& Environmental Sciences (106k).

\textbf{MIMIC-CXR}~\citep{Johnson2019-MIMIC-CXR}
The MIMIC Chest-X-ray was selected as the test bed for our medical communication experiments. 
The original dataset concerns more than 220k radiographic studies performed at the Beth Israel Deaconess Medical Center in Boston, MA.
Even though the dataset is multi-modal, we restrict our experiments to the free-form text reports associated with each radiographic study.
Specifically, all our experiments concern the findings section of each report and no other type of information. 

Given the higher prevalence of certain texts over uncertain texts, we conduct a heuristic approach to pinpoint sentences in the radiology reports that have a few epistemic markers. 
We proceeded as follows: 
    first, we create a whitelist of possible epistemic markers used in radiology reports. These include commonly used expressions (\eg ``possibly'', ``might'', ``perhaps'', ``potentially''), as well as more domain-specific (\eg ``cannot exclude'', ``suspicious for'', ``favored'', ``highly suggestive of'', ``most consistent with'').
    Second, we select all reports whose FINDINGS section contains at least one of the words in the list, discarding all others. 
    Third, we approximate the certainty score of each of the sentences in the resulting files using a BERT model~\citep{wuehrl-etal-2024-understanding}, as this is more scalable than using an LLM-as-a-judge approach. It also seems to be no worse than an average human in terms of correlating with human-level perceptions of certainty. 
    From this dataset, we create two subsets 
        (1) \textit{certain}, which we create by selecting all unique FINDINGS for which there's at least one sentence whose predicted label by the BERT model is CERTAIN (class 2) and the predicted score is $\geq 0.7$; 
        (2) \textit{uncertain}, which we create by selecting all unique FINDINGS for which the predicted label is UNCERTAIN (class 0) with a predicted score $\geq 0.7$. 

For the sentence-level experiments in the main paper, we randomly sample 400 sentences from the \textit{uncertain} subset and 400 sentences from the \textit{certain} subset, yielding 800 sentences drawn from 433 unique reports. 
For the document-level experiments, we sample 800 unique reports whose FINDINGS sections contain at least one sentence classified as uncertain. This stems from the fact that most sentences in the FINDINGS section tend to be descriptive and sound relatively certain, which could make it more difficult to observe certainty distortion effects.

\begin{table*}[t]
\centering
\small
\begin{tabular}{lccc}
\toprule
\textbf{Metric} & \textbf{SPICED} & \textbf{Academic Papers} & \textbf{MIMIC-CXR} \\
\midrule
Number of Unique Source Documents             & 397                 & 500 & 433 \\
Average Number of Sentences     &  1    ($\pm 0.00$)  & 9.07 ($\pm 3.77$) &  6.33 ($\pm 2.06$) \\
Average Number of Words         & 29.38 ($\pm 11.65$) & 245.56 ($\pm 91.47$) & 71.27 ($\pm 31.70$) \\
\bottomrule
\end{tabular}
\caption{Source-document statistics for the corpora underlying the sentence-level experiments. The MIMIC-CXR sentence-level sample contains 800 sentences (400 certain and 400 uncertain) drawn from 433 unique reports.}
\label{tab:apdx:dataset-stats}
\end{table*}

\section{Additional Results}
\label{apdx:sec:additional-results}

\subsection{Semantic Similarity}
\begin{figure*}[tb]
    \centering
    \includegraphics[width=\linewidth]{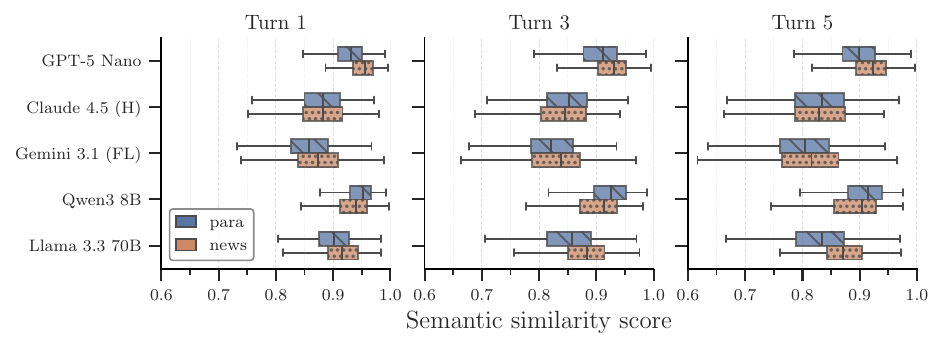}
    \caption{
Semantic similarity scores for all models on the SPICED paraphrase (para) and news-rewrite (news) tasks, broken down by rewriting turn. Scores remain high and relatively stable across turns, with minor variation across models and tasks. These aggregate results indicate that outputs remain semantically similar to their sources on average, even as they exhibit the certainty distortion described in \Cref{sec:experiments}.
}
\label{fig:semantic-similarity:spiced}
\end{figure*}

\begin{figure*}[tb]
    \centering
    \includegraphics[width=\linewidth]{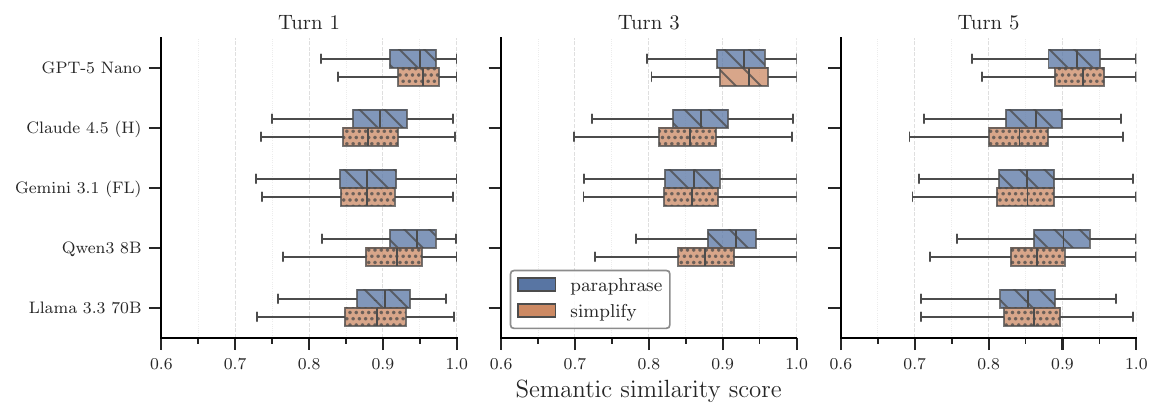}
    \caption{
Semantic similarity scores for all models on the MIMIC-CXR paraphrase (para) and simplification (simplify) tasks, broken down by rewriting turn. Despite minor variation across models and tasks, scores remain high and relatively stable across turns, indicating that outputs remain semantically similar to their source reports on average, even as they exhibit the certainty distortion described in \Cref{sec:experiments}.
}
\label{fig:semantic-similarity:medical}
\end{figure*}

\Cref{fig:semantic-similarity:spiced} illustrates the semantic similarity score per task, model, and LM iteration for  the SPICED dataset. While we observe decreases in semantic similarity with the number of iterations, the scores remain high and we observe no systematic differences between paraphrasing and news tasks, which suggests that the higher \certaintydistortion magnitude observed in the news task (see \Cref{sec:experiments}) is not solely explained by semantic drift.  We observe similar findings in the medical task (see \Cref{fig:semantic-similarity:medical}).

\textbf{Implementation Details.} Semantic embeddings are obtained using the open-source embedding model \texttt{MongoDB/mdbr-leaf-mt} which has shown comparable performance to closed-source embedding models in recent benchmarks~\citep{mdbr_leaf}.

\subsection{Length Effect on Observed Results}
\label{apdx:sec:length-analysis}

{
\Cref{ssec:sentence-level-distortion-rates} shows that models distort the certainty of a non-negligible proportion of examples. 
Communicating uncertainty often requires words or phrases that qualify the core proposition, such as `may', `likely', or `the evidence suggests'. 
When shortening or compressing a text, a model may remove these qualifiers while preserving the core proposition, thereby making the generated text appear more certain.
Hence, it is possible that the observed certainty inflation arises partly as a by-product of this compression, such that tasks producing greater reductions in length exhibit greater inflation. 
We therefore examine whether length differences can fully explain the observed inflation.
}

{
First, we fit a regression predicting certainty inflation while controlling for relative changes in output length, semantic similarity, and model fixed effects. 
Shorter outputs were more likely to inflate certainty ($\beta=-0.96$, $p<0.001$), consistent with the removal of epistemic markers during shortening. 
Nevertheless, the effect of news rewriting remained significant after accounting for these factors ($\beta=0.74$, $p<0.001$). 
These results suggest that, although changes in length contribute to certainty inflation, they do not fully explain the effect observed in news rewriting.
}
\begin{figure*}[tb]
    \centering
    \includegraphics[width=\linewidth]{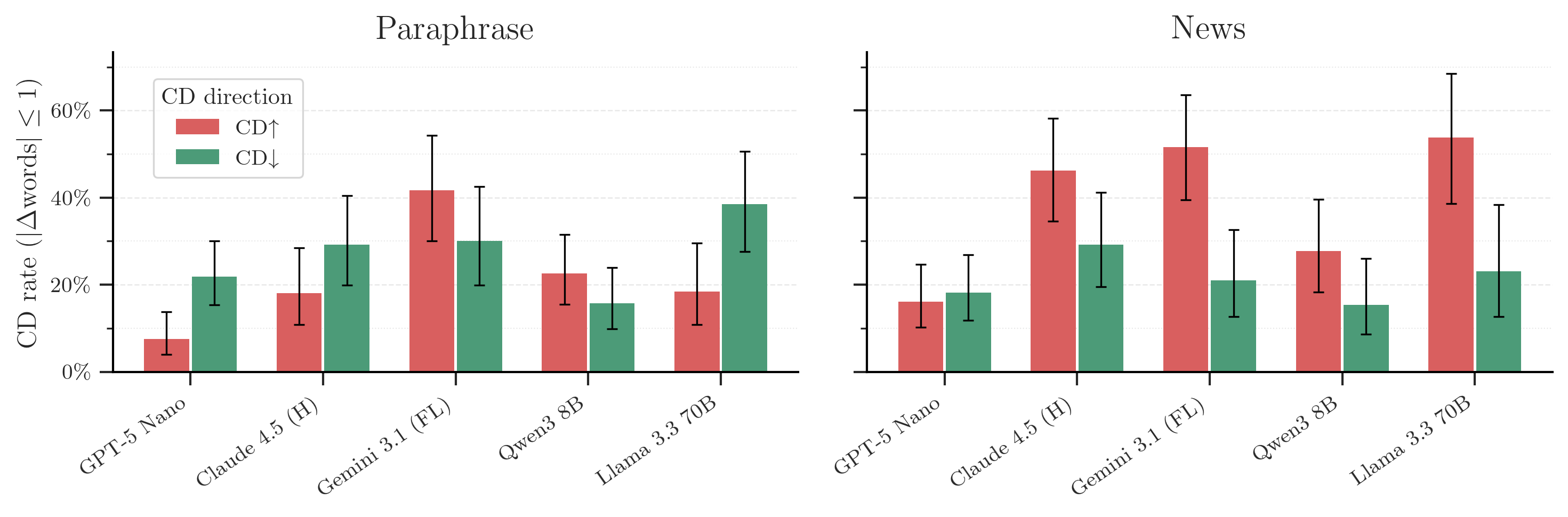} 
    \caption{
    Certainty distortion among approximately equal-length outputs. \certincrease and \certdecrease rates are computed only over examples where the generated output differs from the source by at most one word. Error bars show 95\% confidence intervals. Certainty distortion persists even when output length is nearly preserved, suggesting that length differences contribute to but do not fully explain CD.
    }
    \label{fig:analysis-length-matched-spiced}
\end{figure*}
{
As a second validation, we also conduct a matched-length analysis as a more direct robustness check that does not depend on any statistical model. 
We restricted this analysis to examples for which the generated text differed from the source by at most one word, allowing us to examine certainty distortion when word count was nearly unchanged. 
Certainty distortion persisted in this subset, with its direction varying substantially across models and tasks. 
In particular, approximately 45--55\% of the outputs from several models exhibited increased certainty in the news-rewriting task despite having nearly the same length as their sources (see \Cref{fig:analysis-length-matched-spiced}). 
Together, these results indicate that removing epistemic markers through shortening is one source of certainty distortion, but not its sole explanation.
}

\subsection{Diagnostic Comparison with Human-Written Science News}
\label{ssec:human-spiced-analysis}

{
The higher distortion observed in news rewriting may partly reflect genre conventions learned from human-written science journalism, where hedges are sometimes removed and findings are presented using stronger language. 
To provide a diagnostic comparison, we apply the same CD evaluator used in \Cref{ssec:sentence-level-distortion-rates} to the naturally occurring pairs of academic findings and human-written news counterparts in SPICED. 
The human-written news texts exhibit a certainty distortion rate of 23.9\%. 
By comparison, LM-generated news rewrites exhibit distortion rates ranging from 44.3\% to 75.3\% across models. Thus, the magnitude of LM distortion is higher than the average distortion detected in the corresponding human-written news texts.

We treat this result as a magnitude check rather than a controlled human--LM comparison. The human texts were produced under different conditions and may reflect additional context, editorial objectives, or other genre conventions. 
A matched study in which human writers and LMs receive the same source claims and instructions remains for future work. 
Nevertheless, reproducing certainty shifts learned from human-written news is not epistemically faithful to the source: rewriting should preserve both its semantic content and expressed certainty. Moreover, distortion in scientific paraphrasing and medical communication shows that the phenomenon is not limited to news rewriting.
}

\subsection{Decoding Effect}
\label{apdx:additional-decoding-results}

\begin{figure}[tb]
    \centering
    \begin{subfigure}[b]{\linewidth}
        \centering
        \includegraphics[width=\linewidth]{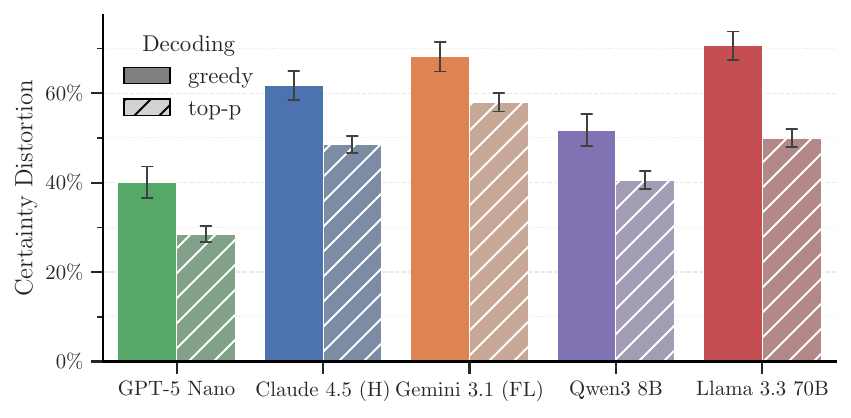}
        \caption{Certainty Distortion Measurements}
        \label{fig:certdistortion-value-differences}
    \end{subfigure}
    \hfill
    \begin{subfigure}[b]{\linewidth}
        \centering
        \includegraphics[width=\linewidth]{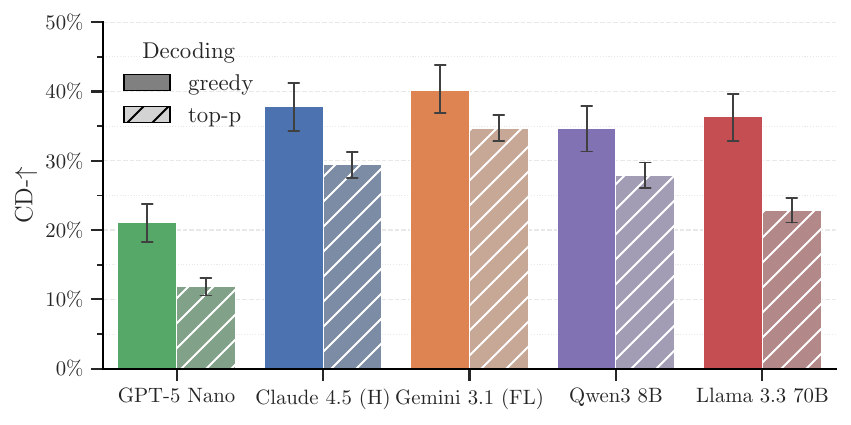}
        \caption{Certainty Inflation Measurements}
        \label{fig:certincrease-value-differences}
    \end{subfigure}
    \hfill
    \begin{subfigure}[b]{\linewidth}
        \centering
        \includegraphics[width=\linewidth]{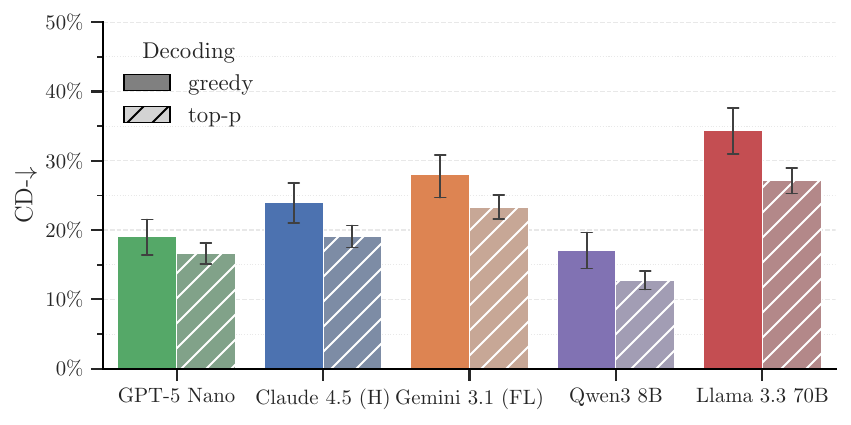}
        \caption{Certainty Deflation Measurements}
        \label{fig:certdecrease-value-differences}
    \end{subfigure}
    \caption{
    Impact of decoding in different certainty measurements, averaged across tasks in SPICED. Greedy decoding implies LM outputs were generated using \texttt{T=0} and top-p decoding concerns the generations with \texttt{top-p=0.9} and \texttt{T=1}.
    }
    \label{fig:decoding:breakdown:spiced}
\end{figure}

\Cref{fig:decoding:breakdown:spiced} shows the results obtained for different models when averaged across news and paraphrasing tasks in SPICED.

{
\subsection{Source Hedge Density and Iterative Distortion}
\label{apdx:broken-telephone-analysis}

To investigate one possible contributor to the different broken-telephone
trajectories in \Cref{ssec:sentence-level-iteration-certainty-distortion-rates}, we conducted a descriptive
analysis of epistemic-marker density in the source texts. Using a
hand-crafted lexicon (the full list is available in the repository), we counted markers in deduplicated source instances
and normalized counts by text length. 
The lexicon comprises numerous modal, evidential, and uncertainty markers, including may, might, could, possible, suggest, and appear; the full list is provided in the repository.
MIMIC sources contained 4.84 markers
per 100 words, compared with 1.54 in SPICED, a 3.13-fold difference. This
is consistent with medical sources providing more uncertainty-bearing cues
that can be progressively weakened across repeated paraphrases. For example,
we observe the following gradual shifts:

\begin{quote}
\textit{``There could be tiny pleural effusions''
$\rightarrow$ ``may be present''
$\rightarrow$ ``are present.''}
\end{quote}

\begin{quote}
\textit{``Opacity ... could represent an early pneumonia''
$\rightarrow$ ``infiltrates are suggested by''
$\rightarrow$ ``infiltration is evidenced by.''}
\end{quote}

These examples illustrate progressive strengthening of commitment to the
same observation rather than the removal of a single hedge. However, the
corpus-level difference does not isolate hedge density from other domain
differences. We therefore treat this analysis as descriptive evidence
consistent with the hypothesis, rather than as a causal explanation of the
different trajectories.
}

\subsection{Robustness Analysis}
\label{apdx:robustness-to-evaluator}

In the main paper, we provide an extensive validation of our sentence-level LM-as-a-judge approach (see \Cref{sec:experiments}).
To validate that our findings are robust to the choice of the evaluator, we replicate our sentence-level evaluation with the most competitive BERT-based metric~\citep{wuehrl-etal-2024-understanding}. 
\Cref{tab:wurl-model-spiced-news-results,tab:wurl-model-spiced-paraphrase-results,tab:wurl-model-medical-paraphrase-results,tab:wurl-model-medical-simplify-results} summarize these results. While we observe that \certaintydistortion values are lower, we find that LMs still exhibit a non-negligible propensity to shift the certainty of the source text. Moreover, we observe that the findings regarding task and domain differences hold under a different evaluator (\Cref{ssec:sentence-level-distortion-rates}).

In the main paper, we observed that \texttt{gpt-5.4-mini} exhibited lower distortion compared to other models in the same family. We hypothesized that these results could be explained by self-preference bias, \ie the bias towards selecting or favoring its own outputs. However, we note that even when evaluating the results with a BERT-based model (fine-tuned on data dating back to 2022), we find that the same patterns are observed, suggesting that the lower \certaintydistortion observed is not an artifact of using \evalgpt as an evaluator, but may actually reflect behavioral differences in how the outputs are generated.

\begin{table}[t]
    \centering
    \small
    \begin{tabular}{lrrrrr}
    \toprule
    \textbf{Model} & \textbf{$N$} & \textbf{CD} & \textbf{CD-$\uparrow$} & \textbf{CD-$\downarrow$} & \textbf{Tie} \\
    \midrule
    \gengpt                 & 397 & 26.4 & 11.8 & 14.6 & 73.6 \\
    \genclaude              & 397 & 31.2 & 20.7 & 10.6 & 68.8 \\
    \gengemini              & 397 & 36.0 & 27.2 &  8.8 & 64.0 \\
    \addlinespace
    \genqwen                & 397 & 24.7 & 18.4 &  6.3 & 75.3 \\
    \genllama               & 397 & 33.2 & 21.7 & 11.6 & 66.8 \\
    \bottomrule
\end{tabular}
    \caption{Certainty distortion in the SPICED paraphrase task, measured using the reference-less fine-tuned BERT-based model proposed by~\citet{wuehrl-etal-2024-understanding}. $N$ denotes the number of examples. CD-$\uparrow$ and CD-$\downarrow$ denote increases and decreases in expressed certainty, respectively. All values except $N$ are reported as percentages.}
    \label{tab:wurl-model-spiced-paraphrase-results}
\end{table}

\begin{table}[t]
    \centering
    \small
    \begin{tabular}{lrrrrr}
    \toprule
    \textbf{Model} & \textbf{$N$} & \textbf{CD} & \textbf{CD-$\uparrow$} & \textbf{CD-$\downarrow$} & \textbf{Tie} \\
    \midrule
    \gengpt              & 397 & 28.7 & 16.1 & 12.6 & 71.3 \\
    \genclaude          & 397 & 38.8 & 24.4 & 14.4 & 61.2 \\
    \gengemini     & 397 & 41.8 & 23.7 & 18.1 & 58.2 \\
    \texttt{gpt-5.4-mini}    & 397 &  27.2 &  10.8 & 16.4 & 72.8 \\
    \genqwen                 & 396 & 32.8 & 20.5 & 12.4 & 67.2 \\
    \genllama  & 397 & 39.3 & 24.9 & 14.4 & 60.7 \\
    \bottomrule
\end{tabular}
    \caption{Certainty distortion in the SPICED news rewriting task, measured using the reference-less fine-tuned BERT-based model proposed by~\citet{wuehrl-etal-2024-understanding}. $N$ denotes the number of examples. CD-$\uparrow$ and CD-$\downarrow$ denote increases and decreases in expressed certainty, respectively. All values except $N$ are reported as percentages.}
    \label{tab:wurl-model-spiced-news-results}
\end{table}

\begin{table}[t]
    \centering
    \small
    \begin{tabular}{lrrrrr}
    \toprule
    \textbf{Model} & \textbf{$N$} & \textbf{CD} & \textbf{CD-$\uparrow$} & \textbf{CD-$\downarrow$} & \textbf{Tie} \\
    \midrule
    \gengpt                  & 800 &  7.2 &  7.2 & 0.0 & 92.8 \\
    \genclaude               & 800 & 13.0 & 12.8 & 0.2 & 87.0 \\
    \gengemini               & 732 &  6.0 &  5.5 & 0.5 & 94.0 \\
    \texttt{gpt-5.4-mini}    & 800 &  3.9 &  3.6 & 0.2 & 96.1 \\
    \genqwen                 & 799 & 12.6 & 11.0 & 1.6 & 87.4 \\
    \genllama                & 800 & 13.0 & 11.8 & 1.2 & 87.0 \\
    \bottomrule
\end{tabular}
    \caption{Certainty distortion in the MIMIC-CXR paraphrase task, measured using the reference-less fine-tuned BERT-based model proposed by~\citet{wuehrl-etal-2024-understanding}. $N$ denotes the number of examples. CD-$\uparrow$ and CD-$\downarrow$ denote increases and decreases in expressed certainty, respectively. All values except $N$ are reported as percentages.}
    \label{tab:wurl-model-medical-paraphrase-results}
\end{table}

\begin{table}[t]
    \centering
    \small
    \begin{tabular}{lrrrrr}
    \toprule
    \textbf{Model} & \textbf{$N$} & \textbf{CD} & \textbf{CD-$\uparrow$} & \textbf{CD-$\downarrow$} & \textbf{Tie} \\
    \midrule
    \gengpt                     & 800 & 3.1 & 3.0 & 0.1 & 96.9 \\
    \texttt{gpt-5.4-mini}       & 800 & 3.1 & 3.0 & 0.1 & 96.9 \\
    \genclaude                  & 800 & 5.1 & 4.2 & 0.9 & 94.9 \\
    \gengemini                  & 800 & 6.4 & 5.8 & 0.6 & 93.6 \\
    \genqwen                    & 799 & 6.4 & 5.9 & 0.5 & 93.6 \\
    \genllama                   & 800 & 5.8 & 4.4 & 1.4 & 94.2 \\
    \bottomrule
\end{tabular}
    \caption{Certainty distortion in the MIMIC-CXR simplification task, measured using the reference-less fine-tuned BERT-based model proposed by~\citet{wuehrl-etal-2024-understanding}. $N$ denotes the number of examples. CD-$\uparrow$ and CD-$\downarrow$ denote increases and decreases in expressed certainty, respectively. All values except $N$ are reported as percentages.}
    \label{tab:wurl-model-medical-simplify-results}
\end{table}

\subsection{Magnitude and Threshold Sensitivity}
\label{apdx:sec:magnitude-threshold-sensitivity}

{
CD is intended as a prevalence measure: it counts any detected directional shift and therefore does not, by itself, characterize the severity of that shift. 
We retain the full ordinal judge predictions and report their empirical distributions in \Cref{fig:turn-effects:heatmap:all-tasks}, allowing alternative thresholds to be applied.

Under a stricter definition that counts only \textit{Clearly Original} and \textit{Clearly Modified} as distortion, strict-CD in news rewriting ranges from 4.3\% to 15.6\% across models. 
Although substantially lower than the corresponding CD rates, these results show that clear shifts remain present. 
Moreover, four of five models still exhibit more upward than downward clear shifts, with inflation-to-deflation ratios between 1.4 and 1.8. 
Thus, absolute distortion prevalence depends on the selected threshold, while the directional tendency toward certainty inflation remains under this stricter criterion. The appropriate threshold for an acceptable change is ultimately application-dependent.
}

\subsection{Multi-turn Effect}
\label{apdx:ssec:multi-turn-effect}

\begin{figure}[]
    \centering
    \begin{subfigure}[b]{\linewidth}
        \centering
        \includegraphics[width=\linewidth]{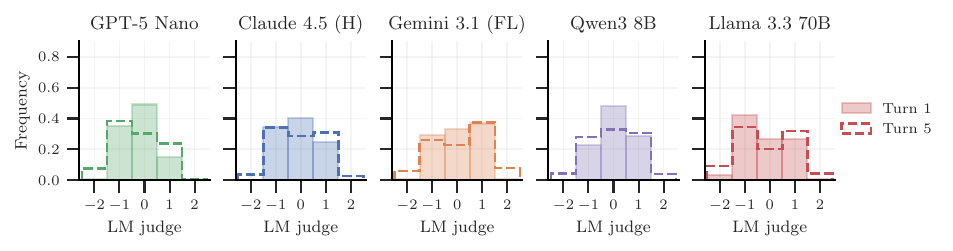}
        \caption{Paraphrase (SPICED)}
        \label{fig:turn-effects:heatmap:all-tasks:para-spiced}
    \end{subfigure}
    \hfill
    \begin{subfigure}[b]{\linewidth}
        \centering
        \includegraphics[width=\linewidth]{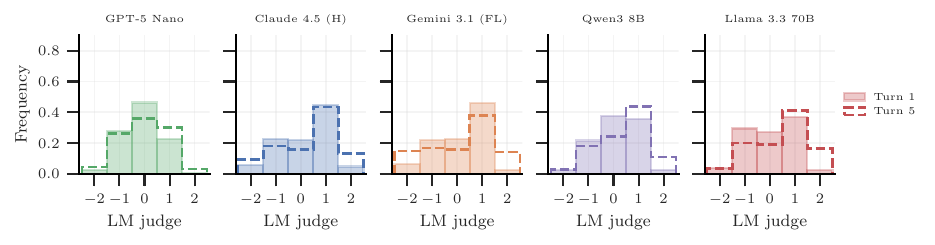}
        \caption{News Rewrite (SPICED)}
        \label{fig:turn-effects:heatmap:all-tasks:news-spiced}
    \end{subfigure}
    \hfill
    \begin{subfigure}[b]{\linewidth}
        \centering
        \includegraphics[width=\linewidth]{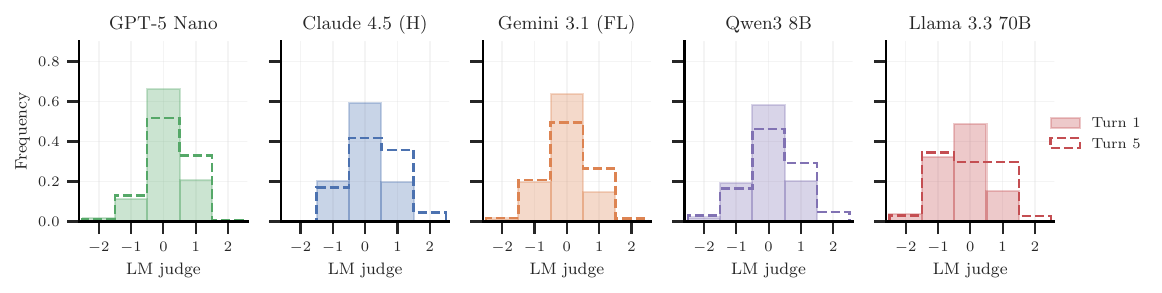}
        \caption{Paraphrase (MIMIC-CXR)}
        \label{fig:turn-effects:heatmap:all-tasks:para-medical}
    \end{subfigure}
    \hfill
    \begin{subfigure}[b]{\linewidth}
        \centering
        \includegraphics[width=\linewidth]{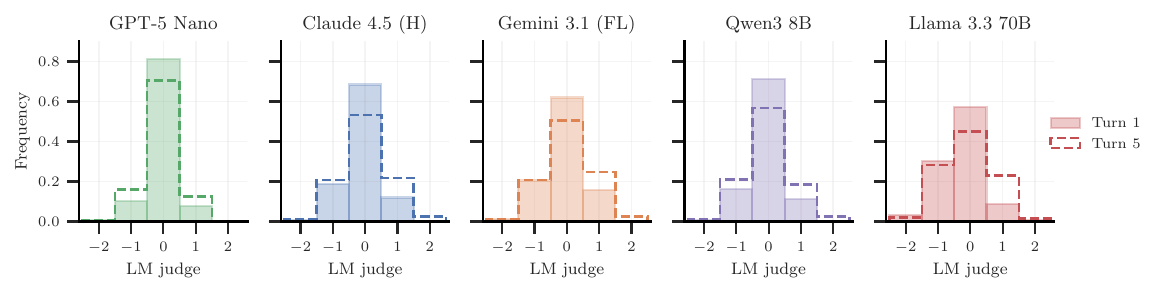}
        \caption{Simplification (MIMIC-CXR)}
        \label{fig:turn-effects:heatmap:all-tasks:simpl-medical}
    \end{subfigure}
    \caption{
    Empirical frequency of LM judge scores across all four task–dataset combinations. Solid bars show scores after a single rewrite (Turn 1); dashed outlines show scores after five successive rewrites of the same source text (Turn 5). Scores range from -2 (source text is more certain) to +2 (model output is more certain), with 0 representing ``No Clear Difference''. Across all conditions, distributions shift rightward with repeated rewriting, indicating that certainty distortion accumulates as the model iterates over its own output rather than the original source.
    }
    \label{fig:turn-effects:heatmap:all-tasks}
\end{figure}

\Cref{fig:turn-effects:heatmap:all-tasks} shows the empirical LM judge score distributions at Turn 1 and Turn 5 for each task and model, illustrating how the distortion profile shifts after five successive rewrites.

\begin{figure*}[]
    \centering
    \begin{subfigure}[b]{0.24\linewidth}
        \centering
        \includegraphics[height=200pt, keepaspectratio]{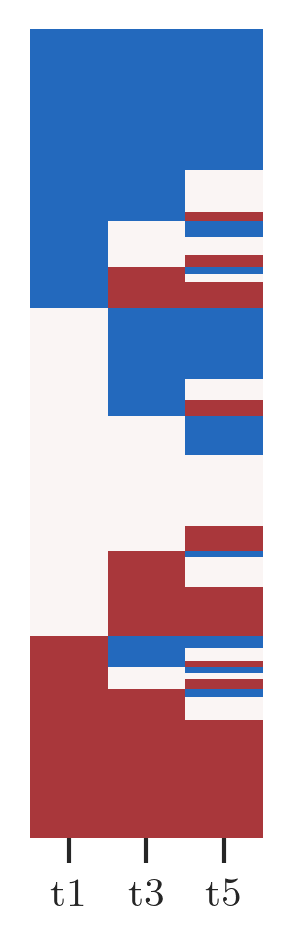}
        \caption{\genclaude \\ Paraphrase (SPICED)}
    \end{subfigure}
    \hfill
    \begin{subfigure}[b]{0.24\linewidth}
        \centering
        \includegraphics[height=200pt, keepaspectratio]{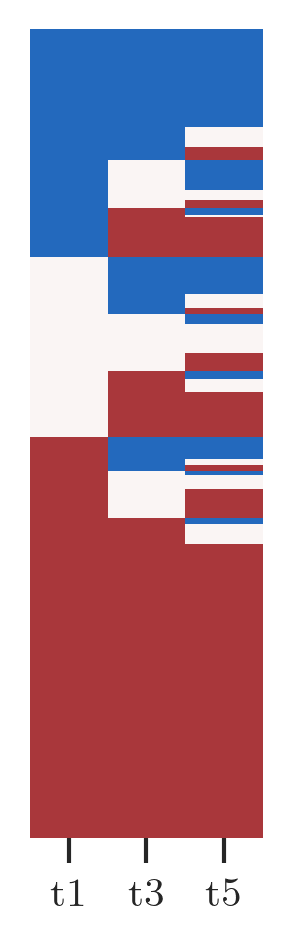}
        \caption{\genclaude \\ News Rewrite (SPICED)}
    \end{subfigure}
    \hspace{1pt}
    \raisebox{3em}{\rule{0.4pt}{200pt}}
    \hspace{1pt}
    \begin{subfigure}[b]{0.24\linewidth}
        \centering
        \includegraphics[height=200pt, keepaspectratio]{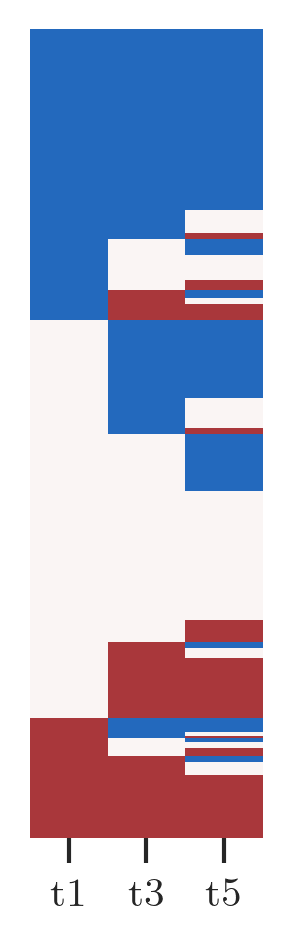}
        \caption{\gengpt \\ Paraphrase (SPICED)}
    \end{subfigure}
    \hfill
    \begin{subfigure}[b]{0.24\linewidth}
        \centering
        \includegraphics[height=200pt, keepaspectratio]{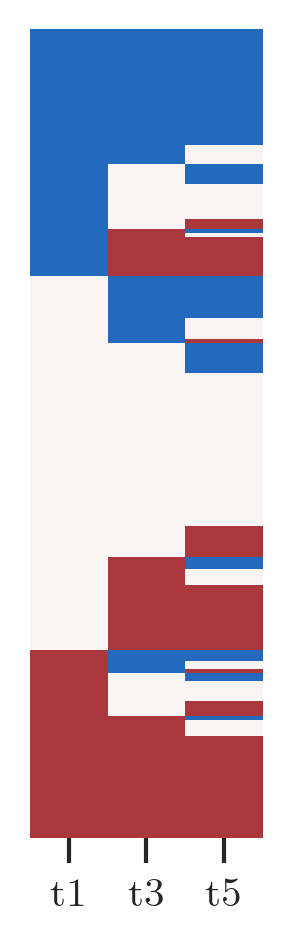}
        \caption{\gengpt \\ News Rewrite (SPICED)}
    \end{subfigure}
    \caption{
    Per-example LM judge score trajectories across successive rewrites for \genclaude and \gengpt on the SPICED paraphrase and news rewrite tasks. \genclaude and \gengpt were selected as models with high and low certainty distortion at Turn 1, respectively. Each row is a source text; blue indicates the source text was judged more certain than the source, red indicates the LM-generated text was judged more certain, and white indicates no clear difference. Rows are sorted by trajectory pattern but are not aligned across panels, each panel represents a different set of source texts.  
    Across all conditions, examples rarely oscillate between red and blue across rewrites; instead, labels tend to stabilize or drift monotonically toward one polarity. \gengpt shows a larger proportion of neutral examples at Turn 1 that subsequently shift toward a definite label, while \genclaude exhibits fewer neutral examples from the outset.
    }
    \label{fig:behavior-heatmaps:spiced}
\end{figure*}

\begin{figure*}[]
    \centering
    \begin{subfigure}[b]{0.24\linewidth}
        \centering
        \includegraphics[height=200pt, keepaspectratio]{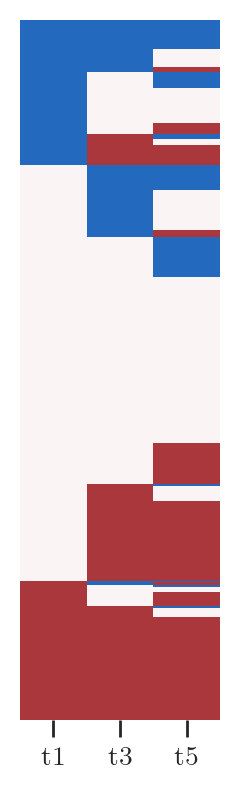}
        \caption{\genclaude \\ Paraphrase (MIMIC-CXR)}
    \end{subfigure}
    \hfill
    \begin{subfigure}[b]{0.24\linewidth}
        \centering
        \includegraphics[height=200pt, keepaspectratio]{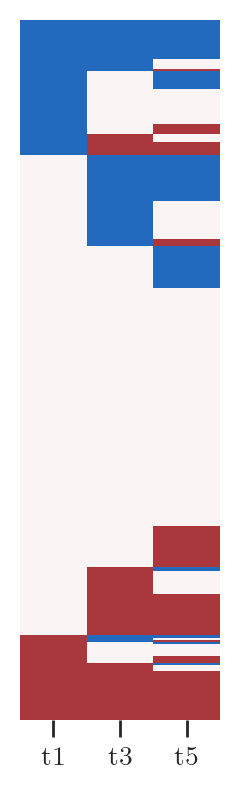}
        \caption{\genclaude \\ Simplification (MIMIC-CXR)}
    \end{subfigure}
    \hspace{1pt}
    \raisebox{3em}{\rule{0.4pt}{200pt}}
    \hspace{1pt}
    \begin{subfigure}[b]{0.24\linewidth}
        \centering
        \includegraphics[height=200pt, keepaspectratio]{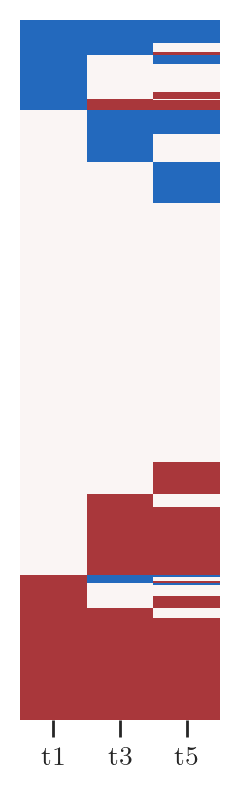}
        \caption{\gengpt \\ Paraphrase (MIMIC-CXR)}
    \end{subfigure}
    \hfill
    \begin{subfigure}[b]{0.24\linewidth}
        \centering
        \includegraphics[height=200pt, keepaspectratio]{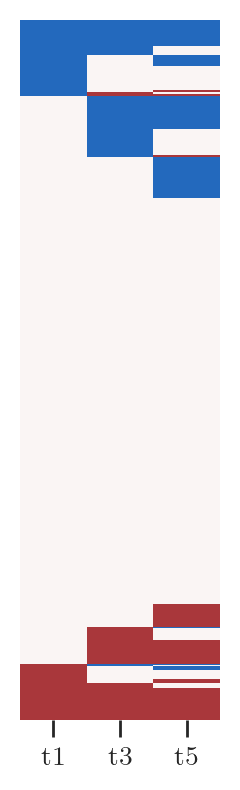}
        \caption{\gengpt \\ Simplification (MIMIC-CXR)}
    \end{subfigure}
    \caption{Per-example LM judge score trajectories across successive rewrites for \genclaude and \gengpt on the MIMIC-CXR paraphrase and simplification tasks. Both models exhibit a directional asymmetry in label transitions: examples that reach a red label rarely revert, while blue labels are comparatively less stable, making blue-to-red transitions more common than the reverse. However, the two models differ in how neutral examples evolve: for \genclaude, the proportion of neutral examples decreases steadily across rewrites, whereas for \gengpt it remains approximately stable.
    }
    \label{fig:behavior-heatmaps:medical}
\end{figure*}

Figure~\ref{fig:behavior-heatmaps:spiced} shows per-example LM judge trajectories across rewrites for models with strong (\genclaude) and weak (\gengpt) distortion trends on SPICED, across both tasks. Each row represents a single source text; colors indicate the LM judge score at that rewrite iteration. Note that rows are not aligned across panels. Each cell corresponds to a different set of source texts, so the comparison is between aggregate trajectory patterns rather than individual examples. Comparing across columns reveals whether drift patterns are task-dependent within a model; comparing across rows shows whether the two models differ in how quickly and consistently distortion accumulates.
~\Cref{fig:behavior-heatmaps:medical} shows the same analysis on MIMIC-CXR. Unlike SPICED, the clinical domain shows a directional asymmetry in label transitions: generations judged as more certain than the source (red) tend to remain so across rewrites, while the reverse is less stable: blue labels (source text more certain) are more likely to transition to red (LM-generated text more certain) than the other way around. The two models differ in how neutral examples (no clear difference between texts) evolve: for \genclaude, the proportion of ties (white) diminishes steadily across rewrites, whereas for \gengpt it remains approximately stable.

\subsection{Additional Qualitative Examples}
\label{apdx:qualitative-examples-}

\Cref{tab:qualitative-examples:orig-certainty-high-models-exacerbate,tab:qualitative-examples:spiced-paraphrase,tab:qualitative-pairwise-examples:random:llms} provide various qualitative examples of LM generations.

\begin{table*}[tb]
\small
\centering
\resizebox{\linewidth}{!}{
\begin{tabular}{p{0.15\linewidth} p{0.42\linewidth} p{0.42\linewidth}}
\toprule
\textbf{Model} & \textbf{Original} & \textbf{LM-generated} \\

    \midrule
\genclaude & There was a \textbf{statistically significant difference} in the global score of SwQoR (SD) between inadequate (37.4 [34.7]) and sufficient FHL (17.7 [21.0]) (P < .001). & Individuals with adequate functional health literacy \textbf{demonstrated substantially superior} sleep quality ratings compared to those with inadequate functional health literacy (P < .001). \\
\rowcolor{gray!10} & Overall, three in four participants \textbf{perceived} the advertisements to breach at least one alcohol-specific rule in the BCAP Code.
    & Three-quarters of study participants \textbf{identified} alcohol advertisements as violating at least one provision of the industry's advertising standards code. \\
\gengemini 
        & Together, we \textbf{provide sound evidence} of a disease-independent unfortunate relationship between well-defined pre-adult environmental hits and violent aggression, \textbf{calling for} more efficient prevention.
        & Our findings \textbf{demonstrate a clear link} between early-life trauma and later violent behavior regardless of underlying health conditions, \textbf{underscoring the urgent need} for improved preventative strategies. \\
\rowcolor{gray!10} 
        & We \textbf{show} that this single meal had an overall and long-lasting impact. 
        & Our findings \textbf{demonstrate} that consuming this one dish resulted in a significant and enduring effect. \\
\bottomrule
\end{tabular}}
\caption{Examples of inflation distortions by \gengemini and \genclaude when the degree of confidence in the source text is high, sampled from SPICED.}
\label{tab:qualitative-examples:orig-certainty-high-models-exacerbate}
\end{table*}

\begin{table*}[tb]
\small
\centering
\resizebox{\linewidth}{!}{
\begin{tabular}{p{0.15\linewidth} p{0.42\linewidth} p{0.42\linewidth}}
\toprule
\textbf{Model} & \textbf{Original} & \textbf{LM-generated} \\
\midrule
\genclaude & The impact of specialty level culture, and peer influence \textbf{appears to} be significant factors of antibiotic decision-making. & The decision to prescribe antibiotics \textbf{is} heavily influenced by the cultural norms within a medical specialty and the opinions of colleagues. \\
\rowcolor{gray!10} & [Conclusion] It was concluded that the 12 weeks of vitamin D supplementation and circuit training \textbf{would have positive effects} on abdominal fat and blood lipid profiles in T2D and vitamin D deficient elderly women.
    & A 12-week regimen involving vitamin D supplementation and circuit training \textbf{significantly reduced} abdominal fat and improved blood lipid profiles in elderly women with T2D and vitamin D deficiency. \\
\gengemini 
        & If a birth is a positive life event, the happiness response to having a child \textbf{may be} strongest when the event is first experienced. 
        & Considering birth as a positive life event, the happiness associated with parenthood \textbf{is} strongest during the first experience. \\
\rowcolor{gray!10} 
        & We \textbf{show} that this single meal had an overall and long-lasting impact. 
        & Our analysis \textbf{proves} that a single meal had a widespread and long-term impact. \\
\bottomrule
\end{tabular}}
\caption{Examples of overconfident distortions in \textit{Paraphrasing} task in SPICED. 
            LMs are prompted to generate paraphrases of an English text that are 1 sentence long.}
\label{tab:qualitative-examples:spiced-paraphrase}
\postspace
\end{table*}

\begin{table*}[tb]
\small
\centering
\resizebox{\linewidth}{!}{
\begin{tabular}{p{0.12\linewidth} p{0.40\linewidth} p{0.40\linewidth} }
\toprule
\textbf{Model} & \textbf{Original} & \textbf{LM-generated}\\
\midrule
 \gengpt    
        & \textbf{These studies provide evidence} that treatment of growth factors can be used to engineer a regeneration response from a non-regenerating amputation wound. 
        & \textbf{The studies provide evidence} that growth-factor treatment can be used to engineer a regenerative response in non-regenerating amputation wounds. \\ 
 \rowcolor{gray!10} \genclaude 
        & \textbf{These studies provide evidence} that treatment of growth factors can be used to engineer a regeneration response from a non-regenerating amputation wound. 
        & \textbf{Scientists have discovered} that growth factor treatment can trigger regenerative healing in amputation wounds that would normally fail to regrow. \\ 
 \gengemini 
        & \textbf{These studies provide evidence} that treatment of growth factors can be used to engineer a regeneration response from a non-regenerating amputation wound. 
        & \textbf{Researchers have discovered} that applying growth factors can coax non-regenerating amputation wounds into regrowing lost tissue. \\
 \rowcolor{gray!10} \genqwen 
        & \textbf{These studies provide evidence} that treatment of growth factors can be used to engineer a regeneration response from a non-regenerating amputation wound. 
        & \textbf{Researchers have discovered that} applying growth factors can induce regeneration in non-regenerating amputation wounds, offering new hope for advanced healing therapies. \\ 
 \genllama  
        & \textbf{These studies provide evidence} that treatment of growth factors can be used to engineer a regeneration response from a non-regenerating amputation wound. 
        & \textbf{Researchers have found} that manipulating growth factors can stimulate regeneration in wounds that would not normally regrow tissue, offering a potential new approach to healing amputations. \\
 \cmidrule{1-3}
  \rowcolor{gray!10}\gengpt
        & The hydrophobic gating mechanism \textbf{is also consistent with} scanning mutagenesis studies \textbf{showing} that modulation of pore hydrophobicity is correlated with activation properties. 
        & Independent scanning mutagenesis experiments \textbf{further support} the hydrophobic gating model, \textbf{showing} that changes in pore hydrophobicity \textbf{track} with the channels’ activation behavior.
        \\ 
  \genclaude 
        & The hydrophobic gating mechanism \textbf{is also consistent with} scanning mutagenesis studies \textbf{showing} that modulation of pore hydrophobicity is correlated with activation properties. 
        & \textbf{Researchers have found} that ion channels' gating behavior is controlled by the hydrophobicity of their pores, a \textbf{discovery confirmed} by genetic studies\textbf{ showing a direct link} between pore water-repelling properties and channel activation. \\ 
  \rowcolor{gray!10}\gengemini 
        & The hydrophobic gating mechanism \textbf{is also consistent with} scanning mutagenesis studies \textbf{showing} that modulation of pore hydrophobicity is correlated with activation properties. 
        & New research \textbf{reinforces} the theory that hydrophobic barriers \textbf{act} as a gate within cellular channels, \textbf{revealing} that altering the water-repelling properties of these pores \textbf{directly dictates} how they open and close. \\
  \genqwen 
        & The hydrophobic gating mechanism \textbf{is also consistent with} scanning mutagenesis studies \textbf{showing} that modulation of pore hydrophobicity is correlated with activation properties. 
        & \textbf{Scientists have found} that the way certain proteins control ion flow \textbf{is linked} to how altering their hydrophobicity affects their activation, as demonstrated by genetic experiments. \\
  \rowcolor{gray!10}\genllama 
        & The hydrophobic gating mechanism \textbf{is also consistent with} scanning mutagenesis studies \textbf{showing} that modulation of pore hydrophobicity is correlated with activation properties. 
        & \textbf{Researchers have found} that the hydrophobic gating mechanism is \textbf{supported} by scanning mutagenesis studies, w\textbf{hich reveal a correlation} between pore hydrophobicity and activation properties. \\
\cmidrule{1-3}
 \gengpt
    & These results \textbf{point to the} \textbf{remarkable} efficacy of our nanoparticle-based drug delivery system and \textbf{suggest} the value of the chicken egg tumor model for testing novel therapies for ovarian cancer.
    & The findings highlight the remarkable efficacy of a nanoparticle-based drug delivery system and suggest that a chicken egg tumor model may provide a valuable platform for testing new ovarian cancer therapies.
    \\
 \rowcolor{gray!10}\genclaude 
    & These results \textbf{point to the} \textbf{remarkable} efficacy of our nanoparticle-based drug delivery system and \textbf{suggest} the value of the chicken egg tumor model for testing novel therapies for ovarian cancer.
    & \textbf{Researchers have} developed a nanoparticle-based drug delivery system that \textbf{shows} \textbf{promise} against ovarian cancer when tested in a novel chicken egg tumor model.
    \\
 \gengemini
    & These results \textbf{point to the} \textbf{remarkable} efficacy of our nanoparticle-based drug delivery system and \textbf{suggest} the value of the chicken egg tumor model for testing novel therapies for ovarian cancer.
    & \textbf{A new study demonstrates} that a novel nanoparticle-based drug delivery system is highly effective in treating ovarian cancer when tested in a chicken egg model, \textbf{offering a promising} new platform for future therapeutic research.
    \\
 \rowcolor{gray!10}\genqwen
    & These results \textbf{point to the} \textbf{remarkable} efficacy of our nanoparticle-based drug delivery system and \textbf{suggest} the value of the chicken egg tumor model for testing novel therapies for ovarian cancer.
    & \textbf{The study highlights} the promising efficacy of a nanoparticle-based drug delivery system for ovarian cancer and underscores the utility of the chicken egg tumor model in evaluating novel therapies.
    \\
\genllama
    & These results \textbf{point to the} \textbf{remarkable} efficacy of our nanoparticle-based drug delivery system and \textbf{suggest} the value of the chicken egg tumor model for testing novel therapies for ovarian cancer.
    & \textbf{Researchers have} made a \textbf{breakthrough} in ovarian cancer treatment, \textbf{demonstrating} the potent effectiveness of a nanoparticle-based drug delivery system in an innovative chicken egg tumor model that \textbf{shows} \textbf{promise} for testing new therapies.
    \\

\bottomrule
\end{tabular}}
\caption{Randomly selected examples from the News Rewriting task. Original texts are randomly selected.}
\label{tab:qualitative-pairwise-examples:random:llms}
\end{table*}

\section{Document-Level Experiments}
\label{apdx:multi-sentence-experiments}

\paragraph{Datasets.} 
Similarly to the sentence-level experiments, we restrict these experiments to the scientific and medical domains.
Specifically, we consider a dataset of 500 academic abstracts, spanning 5 different life and social sciences disciplines, and extend the sentence-level experiments to the full FINDINGS section of 800 MIMIC-CXR reports. Each FINDINGS section in MIMIC was drawn from a pool of uncertain documents, where at least one of the sentences contains an epistemic marker. See Appendix~\ref{apdx:sec:additional-dataset-details} for additional details.

\paragraph{Tasks.} 
We extend the medical tasks to the document-level, prompting LMs to paraphrase or simplify the whole findings section. To facilitate the evaluation under stylistic differences, we further instruct the LMs to preserve the ordering when simplifying the language (see prompt in \Cref{fig:simpli-prompt-medical-doc-level}). 
Similarly, we keep the paraphrasing of full abstracts as one of the document-level tasks and introduce targeted paraphrasing, where we instruct the LM to summarize the main findings in an abstract. With these two tasks, we aim to simulate real-world tasks where users (or authors) may resort to these LMs for help polishing their manuscripts and/or performing a literature review.

\paragraph{LM-as-a-judge Approach.} 
One challenge in expanding the experiments to the multi-document evaluation is to ensure that the LLM-judge bases its certainty judgment decision on semantically equivalent information. In other words, we want to ensure that semantic difference is not biasing our evaluator in unexpected ways. 
To encourage the LM to focus on the common claims across pairs of documents, we tweak the LM-as-a-judge prompt by providing it with examples of expected behavior across the academic and clinical contexts. 

Given a pair of documents (the original document and the modified document), the LM-as-a-judge identifies a list of shared scientific (or radiological) observations for each pair of documents. For each identified claim, the LM-judge has to reason through the sources of certainty used to communicate the claim in both documents and, finally, provide its final output. To differentiate certainty distortion from hallucination, we further instruct the LM (explicitly through a requirement as well as implicitly via example) to produce N/A instead of an ordinal score. To ensure the LM adheres to these constraints, we use Structured Outputs and obtain the response greedily \texttt{T=0}.  The LM-judge is operationalized using \evalgpt. 

Note that due to the implicit nature of the medical data, we found it difficult to find a clear definition of a claim, especially given that the LM does not have full access to the radiological report and supporting images. As a result, in the medical communication tasks, our LM-as-judge compares each sentence in the modified text against the original text (\ie the semantic unit is now a sentence instead of a claim). Other than that, the evaluation process remains the same.

{
\paragraph{LM-Judge Quality Control.} 
To assess the medical document-level evaluator, we constructed a controlled set pairing source sentences with rewrites that preserved, increased, or decreased certainty.
We sampled 20 FINDINGS sections from MIMIC-CXR. From each section, we selected one sentence expressing uncertainty and generated four variants: a paraphrase and three rewrites expressing slightly, moderately, or clearly greater certainty. We repeated this process for one sentence expressing certainty from each section, generating a paraphrase and three rewrites expressing slightly, moderately, or clearly lower certainty.
This process yielded 160 source--variant pairs whose intended certainty direction was known by construction. We used Claude Opus 4.8, Gemini 3 Pro, and GPT-5.4 to generate the variants, increasing stylistic diversity and reducing reliance on artifacts from a single generator.
The first author inspected every variant for clinical plausibility, semantic faithfulness, and consistency with the intended certainty relation; unsuitable or duplicate variants were resampled.
Borderline and extreme cases were additionally discussed with two other authors. This review served as quality control and calibration rather than formal human validation.

We used this controlled set to inspect and refine the two-step LM-judge described above, particularly the alignment step that isolates the shared observation before comparing certainty. The prompt demonstrations cover increased certainty, decreased certainty, hallucinated content, and no clear difference. Qualitative inspection suggested that the evaluator generally followed the intended certainty shifts, particularly for clearer cases, while subtle differences remained ambiguous. Because the controlled set also informed prompt refinement and lacked independent human annotations, we treat this analysis as quality control rather than as formal validation.

This analysis is limited to the medical evaluator. The academic-abstract evaluator uses the same two-step procedure and author-curated demonstrations, but lacks an analogous controlled set; its systematic evaluation remains future work.
}

\paragraph{Metrics.} 
\certaintydistortion, \certincrease, \certdecrease are all computed using micro-averaging, \ie they provide an estimate of how likely a claim is to be distorted.

\paragraph{Prompts.}
The actual prompts used for evaluating the document-level experiments are quite long, so we refer the interested reader to our GitHub repo for the evaluation prompts for \href{https://github.com/UCIDataLab/certainty-distortion-in-llm-rewriting/blob/b4df4538c3da2ee7bb8db4944c63b32dd321599c/configs/experiments/evaluation/prompts/sys-reasoning-5options-academic.txt}{Academic Abstracts} and \href{https://github.com/UCIDataLab/certainty-distortion-in-llm-rewriting/blob/b4df4538c3da2ee7bb8db4944c63b32dd321599c/configs/experiments/evaluation/prompts/sys-reasoning-5options-medical.txt}{Medical} experiments.

\section{AI Usage}

AI assistance was used to polish the writing and assist with styling the plots.
\end{document}